\documentclass[letterpaper, 10 pt, conference]{ieeeconf} 
\IEEEoverridecommandlockouts	
\usepackage{graphicx}
\usepackage{helvet}
\usepackage{authblk}
\usepackage{hyperref}
\usepackage{cite}

\usepackage{amssymb}
\usepackage{amsmath}
\usepackage{commath}
\usepackage{mathtools}
\usepackage{import}
\usepackage{paralist}
\usepackage{svg}
\usepackage{textcomp}
\usepackage{gensymb}
\usepackage{dsfont}
\usepackage{subfig}
\usepackage{graphics}
\usepackage{epsfig} 
\usepackage[normalem]{ulem}
\usepackage[binary-units]{siunitx}
\usepackage{bm}
\usepackage{upgreek}
\usepackage{todonotes}
\usepackage{times}
\usepackage{xurl}
\usepackage{multirow}
\usepackage{makecell}
\usepackage{tabularx}
\usepackage{booktabs}
\usepackage[percent]{overpic}
\usepackage{xcolor}
\usepackage{transparent}
\usepackage{tikzscale}
\usepackage[htt]{hyphenat}
\usepackage{chngcntr}
\usepackage{placeins}
\usepackage{spverbatim}
\usepackage{listings}
\usepackage[capitalise, noabbrev, nameinlink]{cleveref} 
\usepackage{MnSymbol}
\usepackage{subfig}

\usepackage{xspace}
\usepackage[normalem]{ulem}

\usepackage{multicol}
\usepackage{algorithm}

\usepackage{booktabs}
\usepackage{bbm}
\usepackage{wrapfig}
\usepackage[noend]{algpseudocode}
\usepackage{ifthen} 

\definecolor{codegreen}{rgb}{0,0.6,0}
\definecolor{codegray}{rgb}{0.4,0.4,0.4}
\definecolor{codepurple}{rgb}{0.5,0,0.9}
\definecolor{backcolour}{rgb}{0.95,0.95,0.95}
\usepackage{xcolor}

\lstdefinestyle{mystyle}{
    backgroundcolor=\color{backcolour},   
    commentstyle=\color{codegreen},
    keywordstyle=\color{magenta},
    numberstyle=\tiny\color{codegray},
    stringstyle=\color{codepurple},
    basicstyle=\fontsize{10}{10}\selectfont\ttfamily\ttfamily,
    breakatwhitespace=false,         
    breaklines=true,
    breakindent=0pt,
    captionpos=b,                    
    keepspaces=true,                 
    numbers=none,                    
    numbersep=5pt,                  
    showspaces=false,                
    showstringspaces=false,
    showtabs=false,                  
    tabsize=3
}

\newcommand{\citet}[1]{\cite{#1}}

\newboolean{LINK_APPENDIX}
\setboolean{LINK_APPENDIX}{true} 

\def\organa{\textsc{Organa}\xspace}
\def\perception{\textsc{Organa.Perception}\xspace}
\def\nlp{\textsc{Organa.Reasoner}\xspace}
\def\tamp{\textsc{Organa.Planner}\xspace}
\def\analyzer{\textsc{Organa.Analyzer}\xspace}
\def\robot{\textsc{Organa.RobotExecution}\xspace}

\newcommand{\suggest}[2]{{#2}}
\newcommand{\newsuggest}[2]{{#2}}
\newcommand{\revise}[2]{{#2}}
\newcommand{\update}[2]{{#2}}

\algnewcommand{\IfThenElse}[3]{
  \State \algorithmicif\ #1\ \algorithmicthen\ #2\ \algorithmicelse\ #3}
\algnewcommand{\IfThen}[2]{
  \State \algorithmicif\ #1\ \algorithmicthen\ #2}
\algrenewcommand\algorithmicrequire{\textbf{Input:}}
\algrenewcommand\algorithmicensure{\textbf{Output:}}

\DeclareRobustCommand{\SINoteONE}{%
  \ifthenelse{\boolean{LINK_APPENDIX}}
    {\cref{note:supplementary_results:perception-analysis}\xspace} 
    {Note S1\xspace} 
}

\DeclareRobustCommand{\SINoteTWO}{%
  \ifthenelse{\boolean{LINK_APPENDIX}}
    {\cref{note:appendix:nlp}\xspace} 
    {Note S2\xspace} 
}

\DeclareRobustCommand{\SINoteTHREE}{%
  \ifthenelse{\boolean{LINK_APPENDIX}}
    {\cref{note:appendix:nlp:examples-ambiguity-uncertainty}\xspace} 
    {Note S3\xspace} 
}

\DeclareRobustCommand{\SINoteFOUR}{%
  \ifthenelse{\boolean{LINK_APPENDIX}}
    {\cref{note:appendix:user-study}\xspace} 
    {Note S4\xspace} 
}

\DeclareRobustCommand{\SINoteFIVE}{%
  \ifthenelse{\boolean{LINK_APPENDIX}}
    {\cref{note:appendix:hardware-perception}\xspace} 
    {Note S5\xspace} 
}

\DeclareRobustCommand{\SINoteSIX}{%
  \ifthenelse{\boolean{LINK_APPENDIX}}
    {\cref{note:appendix:hardware-skills}\xspace} 
    {Note S6\xspace} 
}

\DeclareRobustCommand{\SINoteSEVEN}{%
  \ifthenelse{\boolean{LINK_APPENDIX}}
    {\cref{note:appendix:parameter-estimaiton}\xspace} 
    {Note S7\xspace} 
}

\DeclareRobustCommand{\SINoteEIGHT}{%
  \ifthenelse{\boolean{LINK_APPENDIX}}
    {\cref{note:appendix:PDDL}\xspace} 
    {Note S8\xspace} 
}
\DeclareRobustCommand{\SINoteNINE}{%
  \ifthenelse{\boolean{LINK_APPENDIX}}
    {\cref{note:appendix:report}\xspace} 
    {Note S9\xspace} 
}
\DeclareRobustCommand{\SIFigPERCEPTION}{%
  \ifthenelse{\boolean{LINK_APPENDIX}}
    {\cref{fig:sample_data_image,fig:perception_pipeline,fig:point_cloud}\xspace} 
    {Figures S1-S3\xspace} 
}

\DeclareRobustCommand{\SIFigTWO}{%
  \ifthenelse{\boolean{LINK_APPENDIX}}
    {\cref{fig:perception_pipeline}\xspace} 
    {Figure S2\xspace} 
}

\DeclareRobustCommand{\SIFigFOUR}{%
  \ifthenelse{\boolean{LINK_APPENDIX}}
    {\cref{fig:ground_perception_gui}\xspace} 
    {Figure S4\xspace} 
}

\DeclareRobustCommand{\SIFigREPORTS}{%
  \ifthenelse{\boolean{LINK_APPENDIX}}
{\cref{fig:report_summ,fig:report_1,fig:report_2,fig:report_3,fig:report_4,fig:report_5,fig:report_6}\xspace} 
    {Figures S5-S11\xspace} 
}

\DeclareRobustCommand{\SITableONE}{%
  \ifthenelse{\boolean{LINK_APPENDIX}}
    {\cref{tab:AP_table}\xspace} 
    {Table S1\xspace} 
}

\DeclareRobustCommand{\SIVideoONE}{%
  \ifthenelse{\boolean{LINK_APPENDIX}}
    {Video S1\xspace} 
    {Video S1\xspace} 
}

\let\oldparagraph\paragraph
\renewcommand{\paragraph}[1]{\oldparagraph*{#1}}

\usepackage{amsmath}

\title{\organa: A Robotic Assistant for Automated Chemistry Experimentation and Characterization}
\author{Kourosh Darvish$^{1,2,3\dagger*}$, Marta Skreta$^{1,2,3\dagger}$, Yuchi Zhao$^{1,2,3\dagger}$,\\Naruki Yoshikawa$^{1,2}$, Sagnik Som$^{1}$, Miroslav Bogdanovic$^{1}$,\\Yang Cao$^{3}$, Han Hao$^{3}$,  Haoping Xu$^{1,2}$,\\ Al\'{a}n Aspuru-Guzik$^{1,2,3\ddagger}$, Animesh Garg$^{1,2,4,5\ddagger}$, Florian Shkurti$^{1,2,3 \ddagger}$ 
}

\begin{document}

\twocolumn[{%
\renewcommand\twocolumn[1][]{#1}%
\begin{center}
    \centering
    \maketitle
    \includegraphics[clip, trim=0cm 1.2cm 0cm 0.8cm, width=0.99\textwidth]{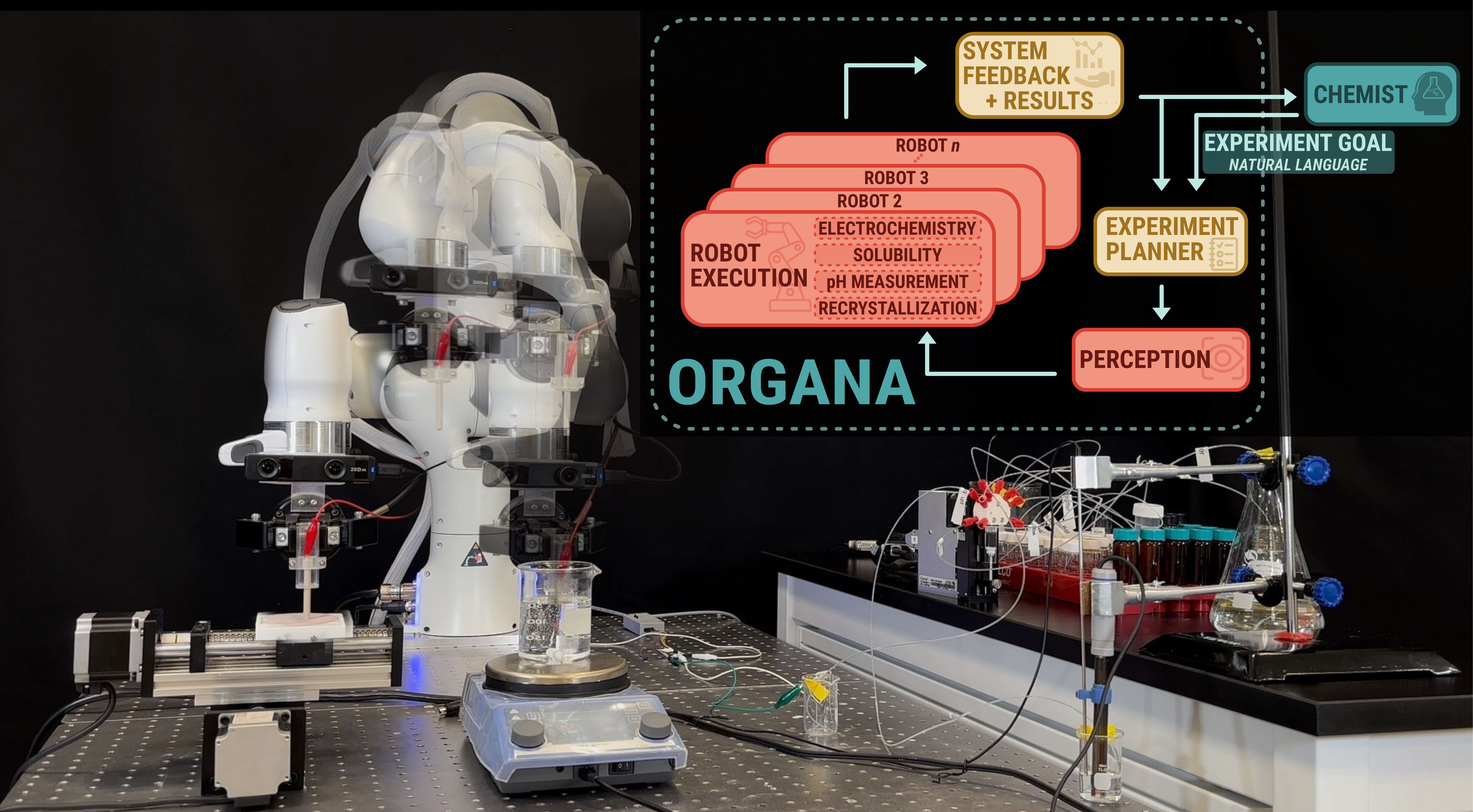}
    \captionof{figure}{\textbf{Robot setup with \organa's overall schema.} \organa provides seamless interaction between SDLs and chemists for diverse chemistry experiments. A key strength of \organa is that it perceives surrounding objects and keeps track of progress on a chemistry task in order to make an informed decision about next steps that are in line with user goals. \organa optimizes SDL efficiency through parallel experiment execution, providing timely feedback via reports and analysis, keeping users well-informed and involved in high-level decision-making. More information about \organa can be found at \href{https://ac-rad.github.io/organa/}{https://ac-rad.github.io/organa/}, including code and a video demonstration.}
    \vspace{15mm}
    \label{fig:fig1_sdl_electrochemistry}
\end{center}
}]

\footnotetext[1]{Department of Computer Science, University of Toronto, Toronto, ON M5S 1A1, Canada}
\footnotetext[2]{Vector Institute, Toronto, ON M5G 0C6, Canada}
\footnotetext[3]{Acceleration Consortium, University of Toronto, Toronto, ON M5G 1X6, Canada}
\footnotetext[4]{NVIDIA, Santa Clara, CA 95051, USA}
\footnotetext[5]{Department of Computer Science, Georgia Institute of Technology, Atlanta, GA 30332, USA}
\renewcommand{\thefootnote}{\fnsymbol{footnote}}
\footnotetext[2]{These authors contributed equally.}
\footnotetext[3]{These authors supervised equally.}
\footnotetext[1]{Lead contact and correspondence: kdarvish@cs.toronto.edu}

\begin{abstract}
Chemistry experiments can be resource- and labor-intensive, often requiring manual tasks like polishing electrodes in electrochemistry. Traditional lab automation infrastructure faces challenges adapting to new experiments. To address this, we introduce \organa, an assistive robotic system that automates diverse chemistry experiments using decision-making and perception tools. It makes decisions with chemists in the loop to control robots and lab devices. \organa interacts with chemists using Large Language Models (LLMs) to derive experiment goals, handle disambiguation, and provide experiment logs. \organa plans and executes complex tasks with visual feedback, while supporting scheduling and parallel task execution. We demonstrate \organa’s capabilities in solubility, pH measurement, recrystallization, and electrochemistry experiments. In electrochemistry, it executes a 19-step plan in parallel to characterize quinone derivatives for flow batteries. Our user study shows \organa reduces frustration and physical demand by over 50\%, with users saving an average of 80.3\% of their time when using it.
\end{abstract}


\section*{INTRODUCTION}
\label{sec:introduction}

The process of discovering materials, from generating candidates to conducting experiments, is time- and labor-intensive. It entails synthesizing and characterizing samples at various scales (ranging from milligrams to grams) in order to identify the desired material that meets the specified requirements. While chemistry labs use a wide variety of special-purpose equipment to expedite independent tasks including synthesis, purification, and analysis, chemists still need to bridge the gaps where automation has not been introduced, such as transferring samples between different workstations. Achieving fully automated Self-Driving Labs (SDLs), with data-driven experiment planning and automated experiment execution, to accelerate material discovery in the vast chemical search space remains a formidable challenge, but one that we are making progress towards \cite{christensen2021automation, roch2018chemos, Mehr2020Universal, burger2020mobile, vescovi2023towards, macleod2022flexible}. This involves integrating general-purpose robots with lab equipment to efficiently perform chemistry tasks, perceive the environment, plan actions, facilitate high-throughput experiments, interact intuitively with chemists, and maintain safety while adapting flexibly to new chemistry tasks.
Furthermore, automatic experimentation in SDLs can enhance result consistency and reproducibility by mitigating the variability inherent in manual experimentation or characterization, which is crucial for scientific discoveries.
Such systems could improve the accessibility of chemistry experiments to users, especially in scenarios of dangerous operations and users with disabilities. 

In this work, we introduce \organa, an assistive robotic solution, as a step toward achieving flexible automation in chemistry labs.
It is a suite of algorithmic tools for robot interaction, perception, and decision making, and is a continuation of our previous work  CLAIRify~\cite{AuRo2023Clairify}.
\organa, as demonstrated in \cref{fig:fig1_sdl_electrochemistry}, uses LLMs to interact with chemists, identify experiment goals, and plan robot experiments. 
It also offers feedback to chemists by analyzing experiment outputs. \organa allows for intuitive communication in natural language with chemists, utilizing either written or speech modalities, thereby reducing human effort and keeping chemists informed of high-level decisions throughout the experiment.
\organa provides chemists with a summary of experiments, their results, and analysis in the form of a report. This approach provides chemists with comprehensive feedback and enables timely user intervention when necessary~\cite{steinruecken2019automatic}.
Another aspect of \organa is its 3D visual perception capabilities, which enables the manipulation of objects as well as monitoring the progress of chemistry experiments. This allows \organa to make informed decisions on how robots will interact with lab equipment and when to proceed to the next step of a long experiment. The combination of autonomous decision-making and high-level human involvement when necessary contributes to the overall robustness of the system, while reducing the amount of manual involvement in the experiment. Additionally, \organa is oriented towards modularity, in terms of both hardware and functional components, empowering scientists to adopt them for various purposes and a diverse set of experiments and hardware setups.
To reduce the overall experiment makespan and enhance efficiency, \organa also supports the parallel execution of chemistry experiments. Readers can refer to \SIVideoONE for a live video demonstration.

As shown in \cref{fig:fig1_sdl_electrochemistry}, \organa receives commands from chemists in audio or text format, translates them using an LLM-based reasoning architecture into an experiment task description, and then maps these instructions to the robot's goals.
Additionally, it grounds perceived objects in the scene through user interaction. \organa improves efficiency by simultaneously solving task and motion planning (TAMP) and scheduling problems, enabling parallel execution of tasks.
Moreover, \organa provides feedback to users by offering a comprehensive report and analysis and notifying them in case of unexpected results during the experiment.
{We show that} \organa {can be} used to execute {four widely-used, fundamental} chemistry experiments: solubility screening, recrystallization, pH experimentation, {and electrochemistry characterization. Solubility screening is an example of using perceptual feedback to make decisions on when to stop the experiment.}
\organa is utilized to identify the electrochemical characteristics of an quinone solution, a promising molecule class for redox flow batteries.
The system is also evaluated by conducting a user study with chemists, validating its \textit{significant usefulness}.
The summary of our contributions is as follows:
\begin{enumerate}
    \item {We introduce a user-friendly assistive robotic system that leans towards a modular design to support fundamental chemistry experiments. \organa reduces the workload of chemists by interacting with them to identify their high-level instructions for experimentation and map them to robot-executable goals. Our system also includes automatic analysis and report generation, as illustrated in \cref{fig:fig2_architecture}.}
    \item {We demonstrate the automation of characterizing the electrochemical properties of a quinone derivative (anthraquinone-2-sulfonate, AQS) through fully automated mechanical polishing, as shown in \cref{fig:fig3_electrochemistry}. Automating mechanical polishing is of significant practical importance to chemists.}
    \item {We solve task and motion planning and scheduling problems together during the planning phase, certifying the execution of tasks by multiple robots and devices in parallel. This extension to TAMP enhances efficiency in experimentation and utilization of lab resources, enabling the parallel execution of chemistry tasks.}
\end{enumerate}

There have been many diverse efforts in lab automation recently; however, many focus on special-purpose hardware and require structured languages~\cite{steiner2019organic, Mehr2020Universal}. Due to its industrial importance, automation of electrochemistry is actively studied~\cite{oh2023electrolab, laws2024autonomous, duke2024expflow}. One of the bottlenecks of these automation systems is the pretreatment of electrodes~\cite{swain2007solid}. Since electrodes get degraded during consecutive measurements, proper treatment is necessary to obtain accurate and consistent results. To address this, Yoshikawa et al.~\cite{yoshikawa2023polish} automated the mechanical polishing of glassy carbon electrodes by combining a custom polishing station and a robot arm, which is a more standard and applicable to more diverse types of electrodes. In this work, we utilize the polishing station proposed by~\citet{yoshikawa2023polish}.

Efforts have been made to overcome the limitation of specialized hardware through the use of a robotic system~\cite{knobbe2022core}.
For example,~\citet{burger2020mobile} employed general-purpose robots to demonstrate the use of a mobile manipulator to operate instruments designed for human chemists and conduct a specific chemistry experiment.
An effort to enhance reconfigurability for chemistry experiments is demonstrated in~\citet{fakhruldeen2022archemist}, employing an intuitive set of state machines and workflows. While these initiatives were conducted in structured lab setups with known object poses, recent works, such as~\citet{vescovi2023towards}, introduce a modular architecture, leveraging QR codes for perception and enhancing flexibility.
Commonly among them, the perception of chemical reactions and objects in the scene ~\cite{LabPic, D3SC05491H, zepel_lai_yunker_hein_2020} introduce significant challenges to lab automation, especially given the transparency of chemistry lab tools \cite{xu2021seeing, wang2023mvtrans}.
{To alleviate robustness and flexibility challenges in lab automation, efforts have been made to use robots as assistive systems.}
{For example, \citet{szymanski2023autonomous} used robotic systems to execute experiments over an extended period with the option of human intervention when necessary.}
In another example, a graphical user interface (GUI) was used in ~\citet{duke2024expflow} to allow users to encode electrochemistry experiments for robot execution using a fill-in-the-blank template.
Despite these promising steps towards SDLs for accelerated material discovery, challenges persist in terms of flexibility, modularity, robustness, and human-centric automation across these examples. 

\begin{figure*}[t]
  \centering

    \includegraphics[clip, trim=1.5cm 0cm 1.5cm 0cm, width=1\textwidth]{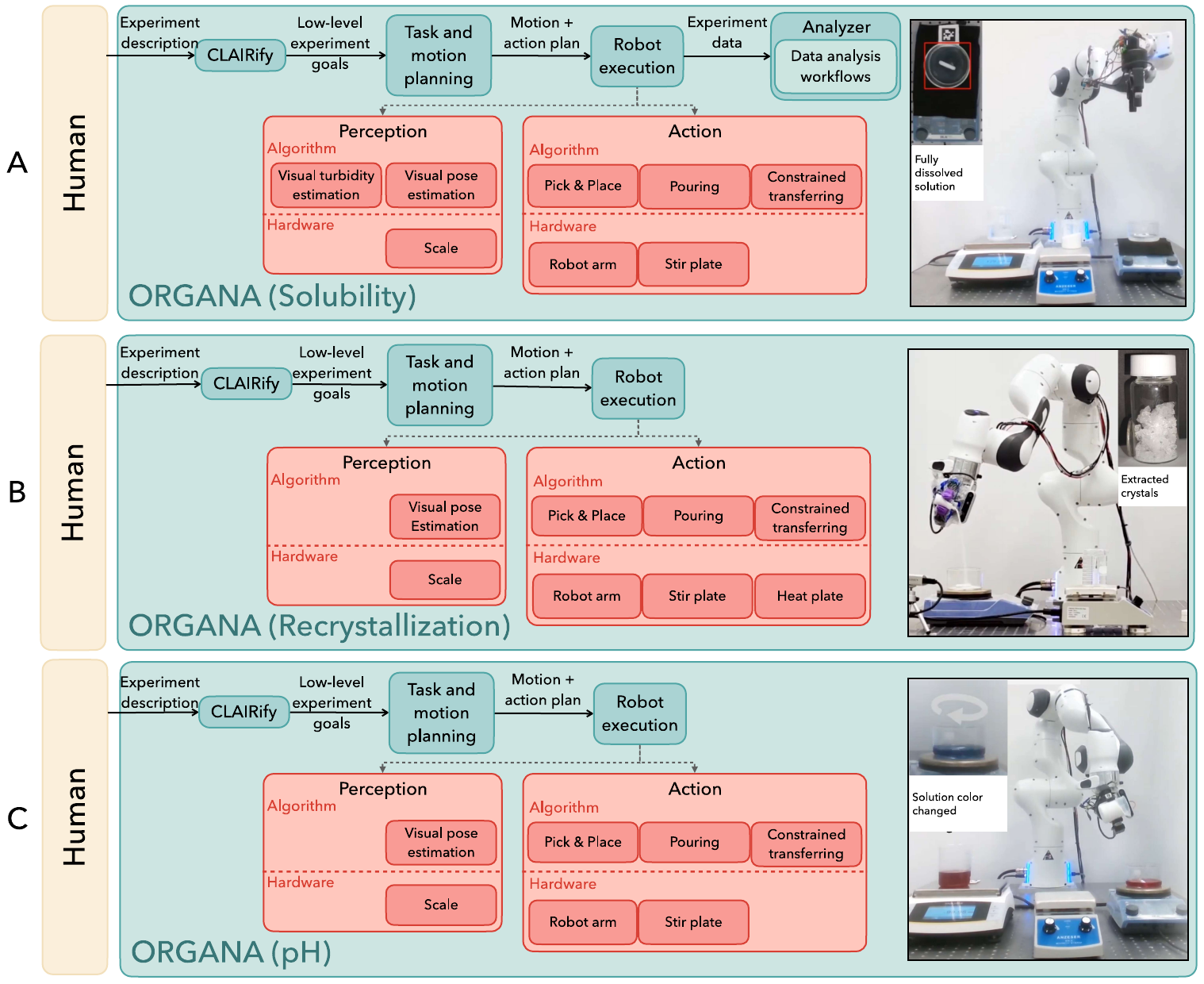}
    \caption{
   \textbf{Instances of \organa conducting various chemistry experiments}: (A) solubility, (B) recrystallization, and (C) pH testing. Images showcase the robot executing actions in each setup, as well as the final results.
    }
    \label{fig:fig4_old_experiments}
\end{figure*}

An inspiring example is found in \citet{aspuru2018matter}
, which envisions a system featuring a fictitious character, Organa, engaging in dialogue with a chemist and providing answers to any chemical question. {In this work, we bring \organa to the real world.} 
{\organa extends our prior work, CLAIRify~\cite{AuRo2023Clairify}, by reasoning over higher-level chemistry tasks through an initial interaction session. It also provides users with feedback about experimental results in the form of reports.
While CLAIRify only reasons over a single experiment, \organa reasons over the chemist's instructions and plans for multiple experiments to run. For example, it may {plan for a series of}  experiments to characterize material properties {over a range of parameters}. Additionally, \organa engages the user in troubleshooting if unexpected behavior, outliers, or ambiguities occur during experiment execution. These additional capabilities of \organa, compared to~\cite{AuRo2023Clairify}, lead to greater efficiency in terms of user engagement and improved robustness in performing experiments.
Moreover, \organa goes beyond previous work in terms of skills, such as the perception of transparent objects, which are essential for executing chemistry tasks without the use of AprilTags~\cite{olson2011apriltag}.
Finally, \organa enhances the efficiency of chemistry experiments by enabling parallel task execution, whereas~\cite{AuRo2023Clairify} only supports sequential execution. This improved efficiency has been achieved by solving TAMP and scheduling problems together, adapting the PDDLStream algorithm~\cite{Garrett2020PDDLStream}. }

Automation in chemistry labs poses a challenge for chemists unfamiliar with robotic programming. Recent efforts have shown that this interaction can be simplified through natural language interfaces.
Chemistry experiment translators have employed rule-based approaches~\cite{Mehr2020Universal} or fine-tuned sequence-to-sequence models for mapping to robot plans~\cite{vaucher2021inferring}. Mehr et al. introduced a hardware-agnostic Chemical Description Language (XDL) in XML format, facilitating integration into diverse robotics infrastructures~\cite{Mehr2020Universal}.

\looseness=-1
Large Language Models (LLMs) {have been shown to be very helpful in } planning chemistry experiments due to their {ability of exhibiting reasoning-like behaviour}~\cite{Ren_Zhang_Tian_Li_2023}, {in that they are able to produce a logical sequence of experiment action steps given their knowledge of the world state}. 
CLAIRify, as an example, utilizes {GPT-3.5} to translate natural language experiments into XDL. {This} process involves iterative prompting of the LLM to ensure syntactic correctness, and the resulting plans can be executed on a Franka robot in a chemistry lab~\cite{AuRo2023Clairify}.
ChemCrow, an LLM-powered chemistry engine, integrates {18} external tools for molecule synthesis planning, including calculators, literature search tools, and IBM’s RoboRXN platform, and successfully demonstrated the synthesis of two real-world molecules from literature~\cite{bran2023chemcrow}.
{Boiko} et al. designed an LLM-based {chemistry agent, Coscientist,} capable of autonomously designing, planning, and conducting scientific experiments by performing tasks like browsing the internet, using liquid transferring robots, and selecting relevant functions from hardware documents~\cite{boiko2023emergent}. 
Inspired by existing approaches that integrate LLM-based models into experiment planning and robot execution, our focus is on monitoring reactions, detecting \textit{unexpected outcomes}, and providing feedback to chemists for corrective actions. 
We {also} explore LLMs generating automatic summary reports for chemists post-experiment, drawing inspiration from the Automatic Statistician project~\cite{steinruecken2019automatic}. Our approach generates reports summarizing robot plans, experiment results, data analysis plots, experiment errors, and human-provided information.

Typically, works in the lab automation literature encode robot and lab device plans using manual state machines\cite{Mehr2020Universal, burger2020mobile}. As a result, they often face challenges in adapting to uncertainties in the environment.
To autonomously solve the problem of robot executable plan generation and parallelizing the robot or agent actions, it is required to solve the problem of TAMP as well as the scheduling problem simultaneously. An approach to solve TAMP {problems} is  {with} PDDLStream~\cite{Garrett2020PDDLStream}, which combines Planning Domain Definition Language (PDDL) solvers for solving discrete planning probelms~\cite{mcdermott1998pddl} with \textit{streams}, a declarative sampling procedure. 
In another work for {a} geometry-rich sequential robot manipulation problem~\cite{toussaint2015logic}, the authors solve it as an optimization problem. 
In an extension,~\citet{toussaint2018differentiable} solves tool-use problems with mixed-integer programming combined with kinematic and differentiable dynamics constraints.
{Learning-based approaches have also been proposed}~\cite{khodeir2023learning, kim2022representation, kumar2023learning} to address the efficiency shortcomings of previous methods.

High-throughput experimentation requires the parallel execution of chemistry tasks, a challenge addressed through scheduling methods discussed in the literature~\cite{hase2019next, hubbs2020deep}.
Scheduling typically deals with a fixed sequence of actions and is commonly resolved using optimization techniques such as mixed-integer linear programming~\cite{longscheduling}.
In this work, we address TAMP and scheduling problems simultaneously.
Simultaneous task planning and scheduling, also known as temporal planning, deals with simultaneous durative actions during planning, often utilizing PDDL2.1~\cite{longscheduling, fox2003pddl2}.
While TAMP and scheduling problems have been explored individually in the literature, their integration has not received sufficient attention.
Two examples of such combinations are presented in~\citet{edelkamp2018integrating}, where temporal planning and a sampling-based motion planner solve the problem in two steps, and in~\citet{Jingkai2021Optimal}, where mixed-integer linear programming was employed.
Furthermore, while~\citet{Garrett2020PDDLStream} provides a simple example code, it lacks a formal description for addressing simultaneous scheduling and TAMP.

\section*{RESULTS}
\label{sec:results}
\looseness=-1
We assess \organa for its reliability in reproducing literature results and modularity through a diverse set of multistep chemistry experiments, 
including solubility screening, recrystallization, pH testing, and electrochemistry. The first three experiments were previously detailed in~\cite{AuRo2023Clairify}. We show them done again here with \organa, and focus on results from an advanced electrochemistry experiment. Additionally, we evaluate the interaction between chemists and \organa through a user study.

\begin{figure*}[t]
    \centering
    \includegraphics[clip, trim=0cm 1.8cm 0cm 0cm, width=0.8\textwidth]{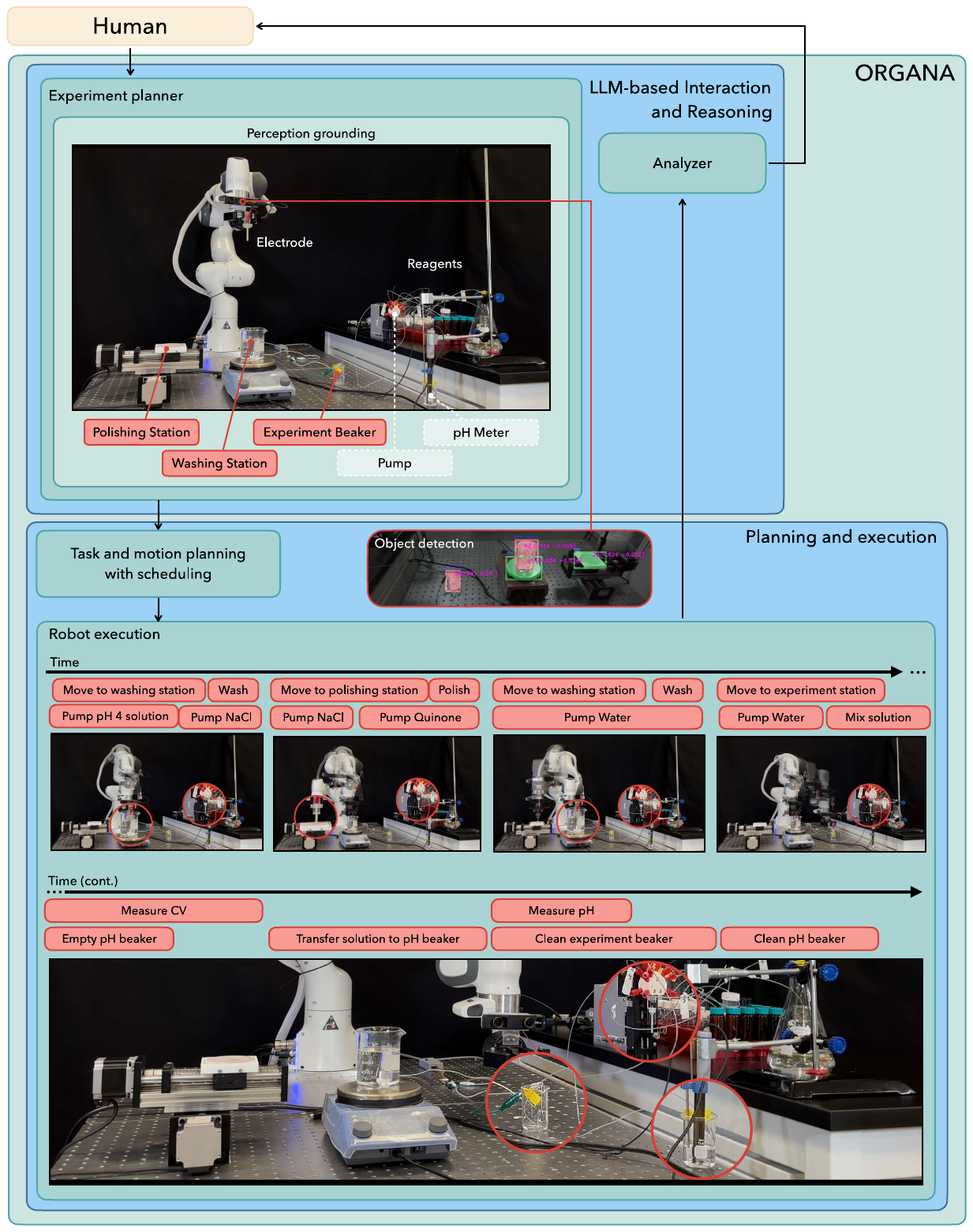}
    \caption{
    \textbf{The electrochemistry setup and experiment workflow.} Initially, users communicate their intention to \organa \suggest{}{via  a text or speech} interface; objects and their poses are perceived. Subsequently, the user interacts with \organa to establish object functionalities. Eventually, \organa plans the robot actions for parallel execution. On top, it shows the experiment setup. At the bottom, it displays the results of visual perception and snapshots of the robot and other hardware executing the actions in parallel.
    }
    \label{fig:fig3_electrochemistry}
\end{figure*}

\subsection*{Solubility Experiment}
\label{sec:solubility}
Solubility is the physical property describing the highest concentration of a solute that is capable of dissolving in a solvent under a specific temperature. 
For this experiment, the robot iteratively pours a small amount of water until all solids are dissolved. After each pouring, the solution is stirred, and the turbidity, a quantitative measurement of residue in solutions, is estimated via a vision-based algorithm that we adapted from~\citet{zepel_lai_yunker_hein_2020}. In \cref{fig:fig4_old_experiments}, an instance of \organa is shown conducting a solubility experiment, with the robot observing the dissolved solution for turbidity value estimation. The system assessed the solubility {(defined as the mass of solute dissolved in 100 g of solvent)} of three solutes in water  —salt (sodium chloride), sugar (sucrose), and alum (aluminum potassium sulfate)— with accuracy values of $7.2\%$, $11.2\%$, and $12.3\%$, respectively, compared to literature results~\cite{nationalhandbook}. The main source of error is attributed to the pouring accuracy of the robot~\cite{AuRo2023Clairify}\suggest{.}{, arising from the delayed response, sensitivity, and limited resolution of the scale and dynamixel motor attached to the robot end-effector.}
For testing each solution, \newsuggest{the system executed a 7-step plan with an average execution time of 25.63 mins.}{the robot and hardware executed a 7-step plan, with each test taking an average of 25.63 minutes.} 

\subsection*{Recrystallization Experiment}
\label{sec:recrystallization}

Recrystallization is a purification process used to extract pure compounds from impure solids. During this process, impure solids are initially added and dissolved in a solution while being heated, and the process is stopped once saturated. As the solution cools, pure compounds start to crystallize due to the solubility of such compounds decreases, while impurities are maintained in the solution. In our experiment, alum was used to test recrystallization because of its dramatic solubility variations in different water temperatures. This experiment modified the solubility test by pre-heating the solvent. \newsuggest{The system}{The robot and hardware} performed a 8-step plan with the execution time of 44.80 mins.
The result of the robot producing crystals is shown in \cref{fig:fig4_old_experiments}, in the middle.

\subsection*{pH Experiment}
\label{sec:ph_experiment}
pH characterizes the acidity or basicity of a solution and is calculated as the negative logarithm of its hydrogen ion activity~\cite{goldbook_ph}.
The anthocyanin pigment in red cabbage can be used as a pH indicator, and the demonstration of its color change is a popular introductory chemistry experiment~\cite{fortman1992demonstrations}.
We prepared red cabbage solution by boiling red cabbage leaves in hot water. The initial color of the solution was dark purple/red.
The color changes to bright pink when an acid is added, and to blue when a base is added.
The robot demonstrated this color change by adding food-grade vinegar (acetic acid, an acid) and baking soda (sodium bicarbonate, a base) into the red cabbage solution.
We applied the pouring skills of liquid and powder to transfer reagents and showcase the resulting color changes in~\cref{fig:fig4_old_experiments}.
To perform this experiment, a 6-step plan was executed with a duration of 3.85 minutes.

\subsection*{Electrochemistry Experiment}
\label{sec:electrochemistry}

\subsubsection*{Task description}
\suggest{Quinone is}{Quinones are}  an important family of molecules that can be applied to metal-free aqueous flow batteries~\cite{huskinson2014metal}, and their electrochemical properties are actively investigated~\cite{khetan2022high}.
The electrochemistry measurements are usually tedious and require human effort for the pretreatment of electrodes which takes several minutes.
To demonstrate the applicability of our robotic system in electrochemistry, we measured the redox potential of a quinone solution at different pH levels and \suggest{drew the}{visualized the corresponding} Pourbaix diagram.
\cref{fig:fig3_electrochemistry} shows our experimental setup.
We \suggest{introduced}{used} a portable, low-cost potentiostat~\cite{garcia2023potentiostat} and a standard 3-electrode system to conduct the electrochemistry experiments.
Notably, we used a glassy-carbon working electrode, which requires mechanical polishing activation, a labor-intensive process in human-centered experiments.
Glassy carbon electrode is one of the most common in electrochemistry studies \suggest{according to a recent survey}{}~\cite{heard2020electrode}.
Although mechanical polishing is the most common method for activating glassy carbon electrodes~\cite {swain2007solid}, it has not been incorporated into existing automation systems. With our robotic system, we introduced a polishing station together with the robotic arm holding the electrode to automate this process~\cite{yoshikawa2023polish}.
    
\looseness=-1
We prepared the quinone solutions at different pHs using a flow-based system based on a syringe pump and selection valves, and the redox potential of the solution was measured using a portable potentiostat described in~\cite{garcia2023potentiostat}.
The solution was a mixture of 2 mM sodium anthraquinone-2-sulfonate (AQS) from Sigma Aldrich, 0.1 M NaCl, and 0.1 M buffer solution.
Six buffer solutions were used for different pH values: acetate buffer (CH$_3$COONa and CH$_3$COOH) for pH 4 and 5, phosphate buffer (Na$_2$HPO$_4$ and NaH$_2$PO$_4$) for pH 6, 7, and 8, and carbonate buffer (Na$_2$CO$_3$ and NaHCO$_3$) for pH 9.
The pH of the buffer solution was adjusted by mixing two solutions manually.
The electrode was mechanically polished for 30~s using a robotic polishing station~\cite{yoshikawa2023polish} to ensure activation and washed in deionized water for 30~s to remove residues.
In each measurement, the open circuit potential (OCP) was measured at the beginning and taken as the starting potential of cyclic voltammetry measurements. Three cycles of cyclic voltammetry measurement were conducted in an electrochemical window between -1.5~V and 0.5~V at a scan rate of 100~mV/s.
The redox potential was calculated by taking the average of oxidation and reduction peak potentials.
The pH of the solution was measured after each electrochemical measurement by automatically transferring the characterized solution to the pH measurement station with the use of a pump.

\begin{figure*}[t]
\centering
\includegraphics[width=0.99\textwidth]{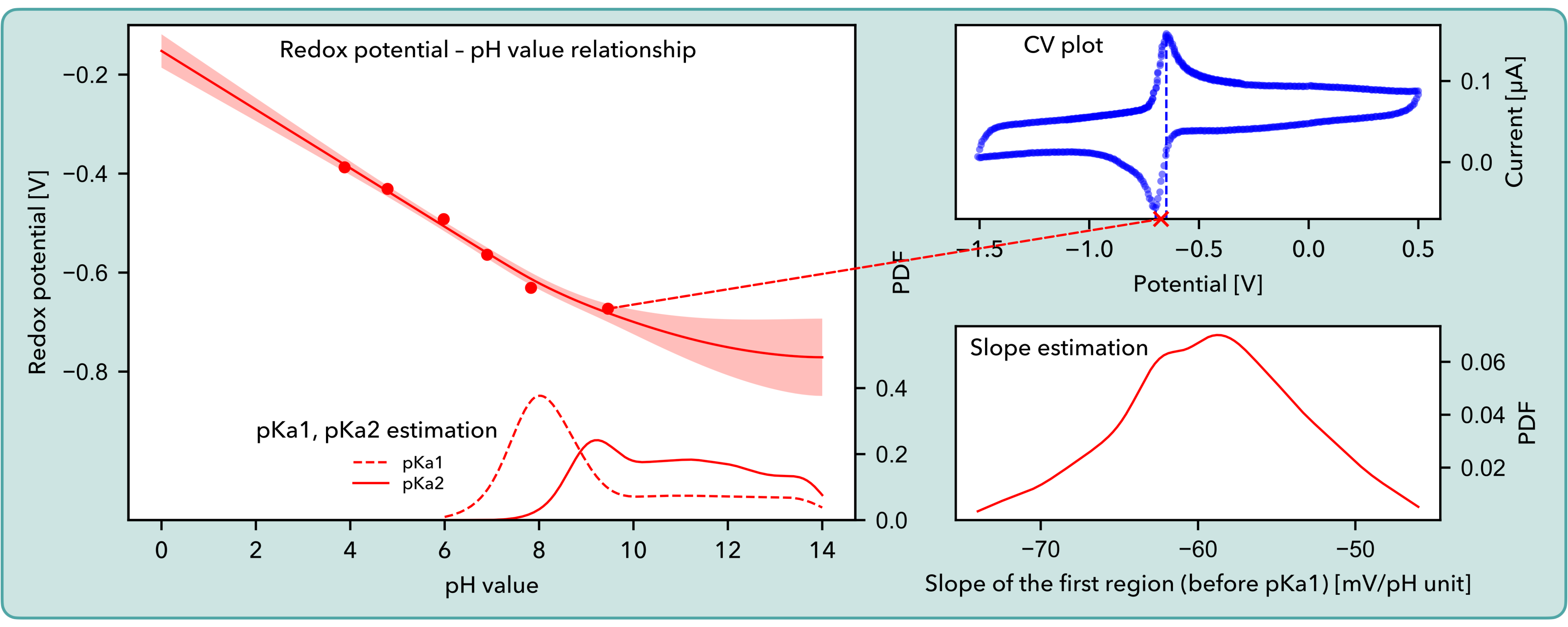} 
\caption{\textbf{The electrochemistry results executed by \organa.} On the left, a Pourbaix diagram is shown for a single \organa experimental run with estimated distributions for p$K_{\mathrm a1}$ and p$K_{\mathrm{a2}}$.
The maximum likelihood estimation (MLE) for p$K_{\mathrm{a1}}$ is 7.86. The top right plot is the cyclic voltammetry curve at pH$=9$. In the bottom right, the estimated slope distribution for the first region is shown, with an MLE of -61.8~mV/pH unit.
The distributions for $pK_{\mathrm{a1}}$, $pK_{\mathrm{a2}}$, and the slope are marginal distributions for individual parameters.
We report the MLE from the full combined posterior distribution over all model parameters in \SINoteNINE, which may be different from the maximum value in the marginal distribution for any given parameter.
}
\vspace{5mm}
\label{fig:electro}
\end{figure*}

\subsubsection*{Experiment setup}
\cref{fig:fig3_electrochemistry} illustrates the setup and workflow employed in conducting electrochemistry experiments. The experimental apparatus \suggest{encompasses various laboratory equipment, including the}{includes a} polishing station, washing station, pH meter, syringe pump, and robot arm. The experimental environment features three distinct beakers: a large one designated for washing, a small vessel for containing the experiment solution, and another small beaker dedicated to pH measurement.
The experimental sequence commences with user interaction with \organa to input experiment details. 
The robot moves toward a view pose, and \perception detects and estimates the poses of objects within the scene (taking approximately 20~s). The user is prompted to ground beakers and stations, specifying the functionality of \suggest{the objects present}{each detected object. This enables the system to understand what each object is used for, which is especially important to consider if there are multiple instances of the same object type}. 
Upon completing these initial steps, \nlp generates a high-level plan and goal, which is fed into \tamp to find a parallel executable plan. 
This process minimizes a cost function associated with total time, thereby maximizing equipment usage.
\robot distributes and executes the plan using multithreading, with each piece of equipment being called asynchronously to perform its respective actions. \cref{fig:fig3_electrochemistry} showcases snapshots of electrochemistry task in progress.

\looseness=-1
Throughout the experiment, pH values and redox potential values are recorded and provided to the \analyzer for estimating peak voltage at each pH level. Subsequently, the \analyzer compares these results with the expected outcomes suggested by \nlp based on the initial information provided by the user. If inconsistencies arise, \organa notifies the user and provides feedback for troubleshooting; otherwise, the experiment proceeds autonomously without user intervention, with \suggest{the LLM}{\nlp}  proposing the next buffer solution and the corresponding high-level plan for execution. After testing all buffer solutions, a summary report, along with analyzed results, is automatically generated for user review. An example of such a report is provided in \SINoteNINE and \SIFigREPORTS.

In total, three complete electrochemistry experiments were performed, testing six buffer solutions for each experiment, ranging from pH 4 to pH 9. Further details regarding the prompts in the human-\organa interaction are available in \SINoteTWO and \SIFigFOUR.

\begin{figure*}[t]
\centering
\includegraphics[width=0.99\textwidth, trim=0 0 0cm 0cm, clip]{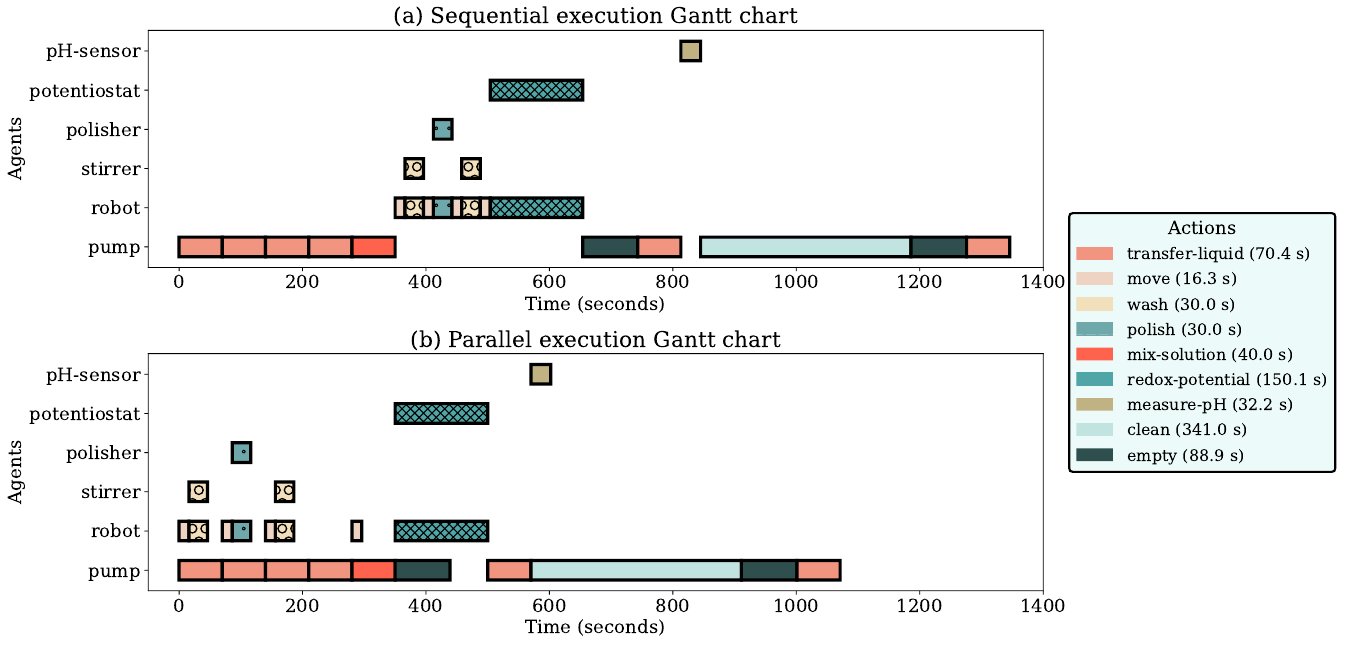}

\caption{\textbf{Gantt chart of the electrochemistry experiment with sequential and parallel execution.} The execution times are written in the legend. \suggest{}{For sequential execution as shown in (a), only one action is executed at a given time and it takes 1,346 seconds (22.43 min) to complete a single test of a buffer solution. For parallel execution shown in (b), the total execution decreases to 1,071 seconds (17.85 min) since multiple actions associated with different agents can be executed at the same time.}  Note: pump cannot transfer solution to pH beaker at $\sim$500 sec \suggest{}{in the parallel execution experiment} due to ongoing redox potential measurement. Boxes stacked on top of each other with the same color and pattern mean that the agents are involved in performing the same action jointly.
    }
\label{fig:gantt-chart}
\end{figure*}

\begin{figure*}[t]
    \centering
    \includegraphics[width=0.95\textwidth]{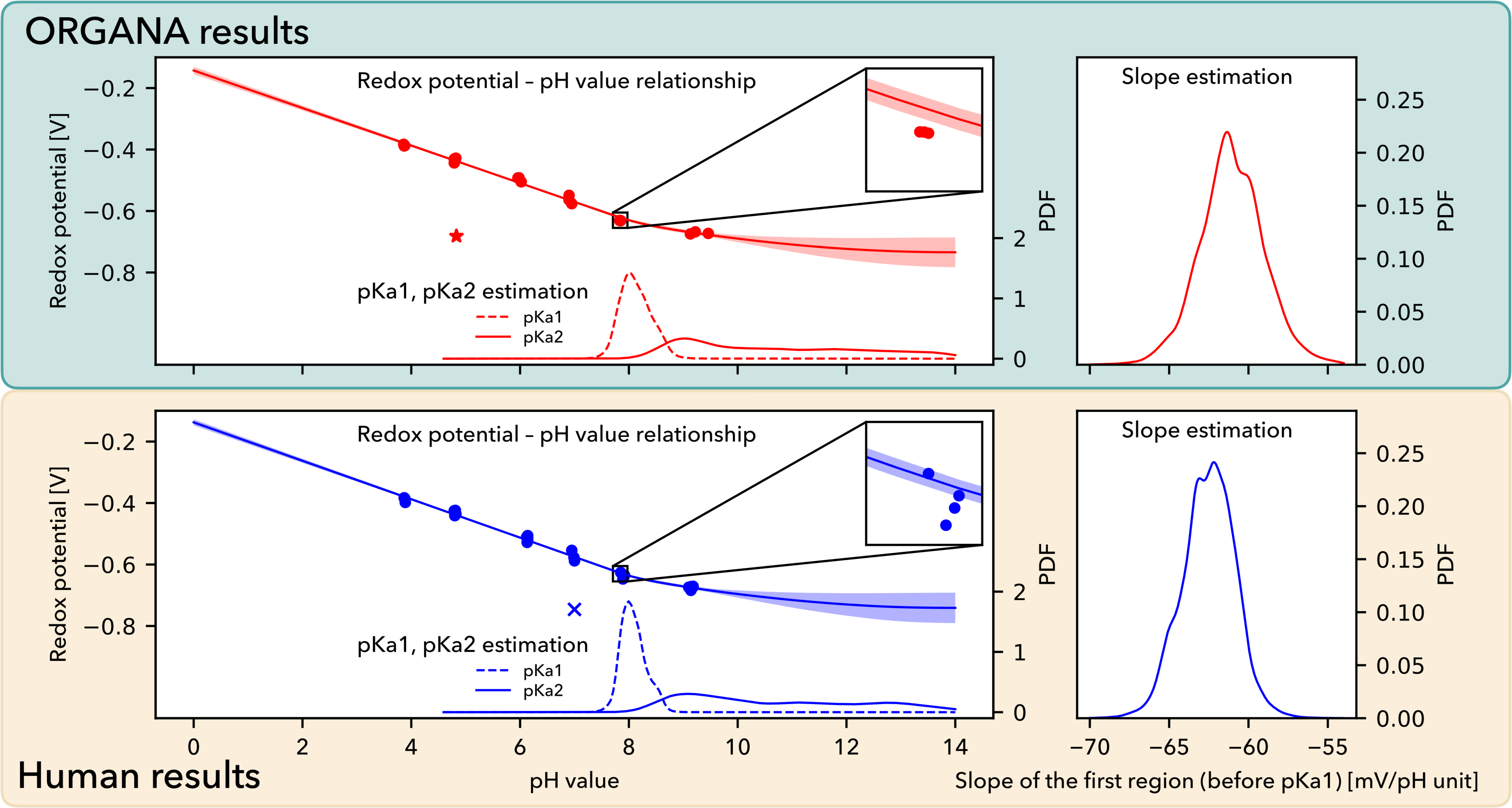} 
    \caption{ \textbf{Comparison between electrochemistry results conducted by chemists and \organa.}
    Comparison of Pourbaix diagrams and their first region estimated slopes in electrochemistry experiments conducted by \organa (top, with 3 data points per pH value) and chemists (bottom, with 4 data points per pH value). Results are comparable: \organa with $pK_{\mathrm a1}$=8.03 and chemists with $pK_{\mathrm a1}$=8.02.
    The estimated slope for \organa is -61.3~mV/pH unit and for chemists is -62.7~mV/pH unit.
    \revise{}{Distributions shown for $pK_{\mathrm a1}$, $pK_{\mathrm a2}$, and slope are marginal distributions for individual parameters. We report the maximum likelihood estimate from the full combined posterior distribution over all model parameters in \SINoteNINE, which may be different from the maximum value in the marginal distribution for any given parameter.}
    The red star and blue cross highlight two distinct problems with task execution.
    }
    \label{fig:pourbaix}
\end{figure*}

\paragraph{Quinone Characterization}
\suggest{Quinones typically undergo three different types of reactions depending on the pH of the solution, and they are shown in Figure~\ref{fig:electro} as the different slopes in the Pourbaix diagram.
Two-proton/two-electron reaction is dominant in the region where pH is smaller than p$K_{\mathrm a1}$, one-proton/two-electron reaction is dominant between p$K_{\mathrm a1}$ and p$K_{\mathrm a2}$, and zero-proton/two-electron reaction is dominant when pH is larger than p$K_{\mathrm a2}$~\cite{khetan2022high}.
The redox potential for $m$ proton, $n$ electron redox couple changes $-59\: m/n$ mV/pH unit at 25~°C, according to the Nernst equation~\cite{quan2007voltammetry}.
As a consequence, the slope of Pourbaix diagram is predicted to be -59~mV/pH unit in the first region where pH is smaller than p$K_{\mathrm a1}$, -30~mV/pH unit in the second region between p$K_{\mathrm a1}$ and p$K_{\mathrm a2}$, and 0~mV/pH unit in the third region where pH is larger than p$K_{\mathrm a2}$.
The reported dissociation constants of AQS are p$K_{\mathrm a1}$ = 7.68 and p$K_{\mathrm a2}$ = 10.92~\cite{quan2007voltammetry}.
Our experimental results agree with these theoretical predictions.
In three repeated experiments the estimated values for the slope of the leftmost region were: -61.3, -61.8 and -61.0~mV/pH~unit.
The estimated values for the dissociation constants p$K_{\mathrm a1}$ were: 8.12, 7.86 and 8.10.
As can be seen in Figure~\ref{fig:electro}, representing the results of a single experimental run, even with only 6 measurements we can achieve a low variance estimate for the slope.
The variance for p$K_{\mathrm a1}$ value is higher, caused by lack of points for higher pH values.
However, even with the given set of pH values, we can produce a lower variance estimate for p$K_{\mathrm a1}$ by utilizing combined data from all three experiments, as can be seen in Figure~\ref{fig:pourbaix}.}{}

\suggest{}{
Quinones typically undergo three different types of reactions depending on the pH of the solution, and they are shown in \cref{fig:pourbaix} as the different slopes in the Pourbaix diagram.
Two-proton/two-electron reaction is dominant in the region where pH $<$ p$K_{\mathrm a1}$, one-proton/two-electron reaction is dominant between p$K_{\mathrm a1}$ and p$K_{\mathrm a2}$, and zero-proton/two-electron reaction is dominant when pH $>$ p$K_{\mathrm a2}$~\cite{khetan2022high}. This results in different slope values on the Pourbaix diagram: -59~mV/pH unit  where pH $<$ p$K_{\mathrm a1}$, -30~mV/pH unit  where p$K_{\mathrm a1}$ $>$ pH and pH $<$ p$K_{\mathrm a2}$, and 0~mV/pH unit  where pH $>$ p$K_{\mathrm a2}$.
The reported dissociation constants of AQS are p$K_{\mathrm a1}$ = 7.68 and p$K_{\mathrm a2}$ = 10.92~\cite{quan2007voltammetry}. Because p$K_{\mathrm a2}$ (10.92) is in a corrosive region of pHs, we only investigated pHs to solve for p$K_{\mathrm a1}$.  Our experimental results agree with these theoretical predictions. \organa obtained the following slope estimates for pH $<$ p$K_{\mathrm a1}$: -61.3, -61.8 and -61.0~mV/pH unit .
The estimated values for the dissociation constants p$K_{\mathrm a1}$ were: 8.12, 7.86 and 8.10. As can be seen in \cref{fig:electro}, representing the results of a single experimental run, even with only 6 measurements we can achieve a low variance estimate for the slope.
The variance for p$K_{\mathrm a1}$ value is comparatively high, caused by a lack of points at pH values greater than 9 because of safety concerns in the robotics lab.
However, even with the given set of pH values, we can reduce p$K_{\mathrm a1}$ estimation variance by utilizing combined data from all three experiments, as can be seen in \cref{fig:pourbaix}.}

\paragraph{Sequential and parallel task execution and efficiency}
We have performed the electrochemistry experiments both sequentially and with parallel task plans. 
\cref{fig:gantt-chart} demonstrates the Gantt chart of the parallel task plan.
The available agents to perform the actions are indicated on the y-axis of the Gantt chart.
When solving the concurrent task and motion planning problem with durative actions, to avoid state space explosion, we assume three possible action durations of 1-3~T  for simplification, where T is the action unit time.
Actions with a duration of up to 60 seconds are simplified as 1T, 2T (up to 120 seconds), and 3T (more than 180 seconds).
In total, the plan is a sequence of 19 actions, some performed by a single agent, while others are performed as a joint action with several agents being involved.

On average, the sequential plan takes 21.67 mins to execute, while the parallel plan takes 17.10 mins to execute, reducing the total time significantly by 21.1\%.
Finally, the averages and standard deviations of the planning time over 12 trials for the above cases are: solving the sequential planning problem ($61.52\pm0.1$ s); and temporal task and motion planning with the time-variant cost function associated with total time ($186.3\pm46.0$ s).

\subsection*{User Study} \label{sec:human_study}

To assess the usability of our system in a real-world laboratory setting, we conducted a study with experimental chemists. The details of participants in this study are provided in \SINoteFOUR.

\looseness=-1
\subsubsection*{Test modes}
We asked subjects to perform the following tests:

\begin{enumerate}
 
\looseness=-1
\item \textit{Manual experimentation}:  Chemists manually conducted an electrochemistry experiment to generate a Pourbaix plot. Following the procedure outlined in  \SINoteFOUR, they prepared a quinone solution using pipettes and manually polished an electrode. Following this, they conducted a CV scan and measured the solution's pH with the same potentiostat and pH sensor used by the robot. Each human subject performed measurements at three different pH values, leading to a total of 24 data points, i.e., four measurements per pH value.

\item \textit{\organa startup}: Users interact with \organa at the startup phase to provide their intention of the chemistry experiment, i.e., information on experiment procedures, goals, expected observations, and vessel semantics. Chemists were asked to follow a script with the target Pourbaix experiment, first by writing, and then by speaking.

\item \textit{\organa troubleshooting}: Users were asked to interact after obtaining information from the start-up phase. To test user engagement, a scenario was designed where an intentional nonsensical observation triggered \nlp's rationalization component to prompt user feedback, ensuring active involvement despite the system's potential autonomous execution capability.

\item \textit{CLAIRify}: Users were instructed to interact with \organa as if the language planner module did not exist, manually detailing each experiment step. This is the equivalent of using CLAIRify~\cite{AuRo2023Clairify}. 

\end{enumerate}

\subsubsection*{Evaluation metrics}
We assessed both quantitative and qualitative metrics~\cite{darvish2023teleoperation}, including user interaction time during experiment execution and the variance in Pourbaix plot measurements. Qualitative analysis involved three surveys.

\begin{enumerate}

\item \textit{NASA Task Load Index (NASA-TLX)}: It assesses a participant's perceived workload across mental demand, physical demand, temporal demand, effort, performance, and frustration level~\cite{hart2006nasa}. Participants self-rated on a scale of 0 (low, good) to 20 (high, bad). NASA-TLX was administered after both manual experimentation \suggest{(\textbf{T1})}{} and \organa usage, with the questionnaire details in  \SINoteFOUR.

\item \textit{System Usability Survey (SUS)}: It evaluates subjective usability with ten Likert scale questions~\cite{sus}. Post-interview, scores range from 0 to 10, with higher scores indicating better usability. Users completed the SUS after both manual experimentation \suggest{(\textbf{T1})}{} and interacting with \organa \suggest{(\textbf{T2} and \textbf{T3})}{}. The survey questions are detailed in \SINoteFOUR.

\item \textit{Custom questionnaire}: We created a 24-question custom Likert scale questionnaire to gauge user preferences for various system components and perceptions on lab experiment automation. 
Questions are framed positively and negatively to alleviate response bias among subjects~\cite{furnham1986response}. \revise{The questionnaire is presented in \cref{fig:human-subjective-study}.}{We show these questions and response summaries in \cref{fig:human-subjective-study}; to visualize preferences more easily, we inverted the scores of negative questions. The original survey can be viewed in \SINoteFOUR.
}
\end{enumerate}

We performed a one-tailed T-score evaluation at a significance level of $p < 0.05$ for the SUS and NASA-TLX studies to gauge differences between the manual experiment and \organa.

\begin{figure*}[t]
    \centering
    \includegraphics[width=\textwidth]{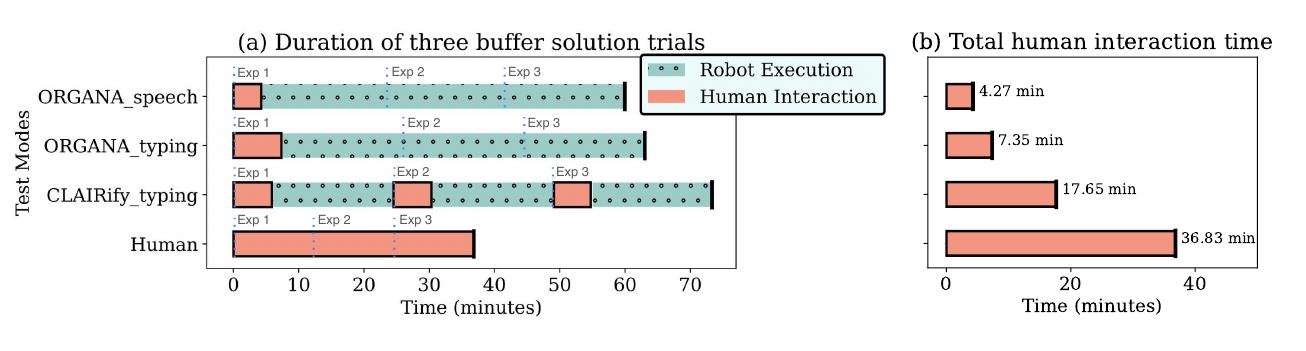}
     
    \caption{
    \textbf{Chemist interaction time with \organa in various test modes. \suggest{(\textbf{T1}-\textbf{T4}).}{}} As shown in (A), the x-axis represents the duration for three iterations of the buffer solution trials in the electrochemistry experiment. This figure shows the timing and frequency of human-\organa interactions, with total times specified in \revise{}{(B)}. \revise{Intermediate user interactions (highlighted with hatched boxes) in \organa modes arise from interviewer-introduced troubleshooting, while other modes require user interaction for task execution.}{It is clear that \organa requires less frequent user intervention compared to CLAIRify and manual experimentation. } \suggest{}{(B) demonstrates the total human interaction time in each test mode \revise{It is clear that both vocal and textual \organa interaction modes are more efficient compared to manual experimentation and CLAIRify}{, showing that both vocal and textual ORGANA interaction modes are more efficient compared to manual experimentation and CLAIRify}. }
    }
    \label{fig:interaction_time}
\end{figure*}

\begin{figure*}[t]
  \centering

    \includegraphics[width=0.8\textwidth]{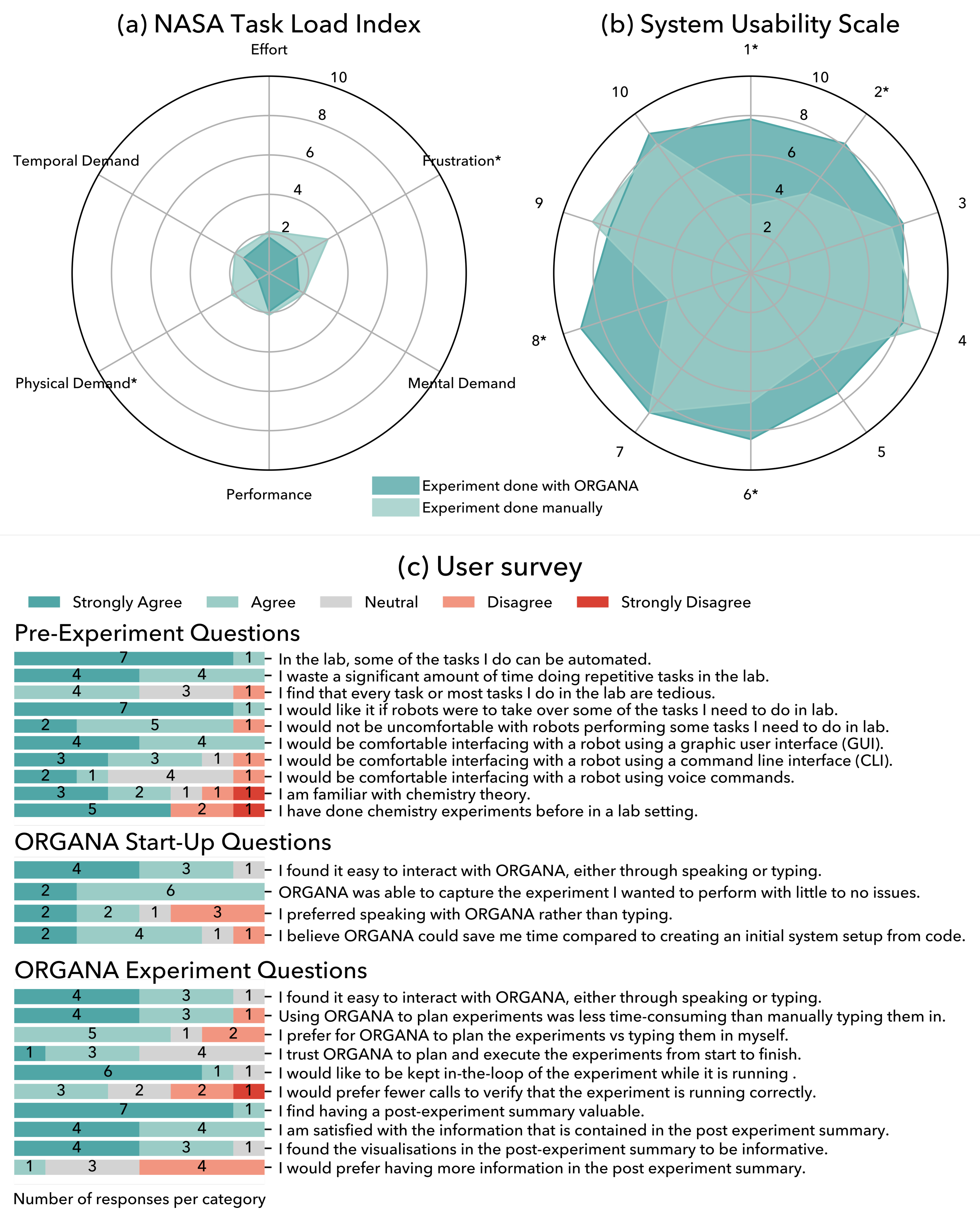}
   
  \caption{\textbf{Human subjective evaluation.} (A) NASA Task Load Index (the lower, the better). (B) System Usability Scale (the higher, the better). (C) User study questions.
  Asterisks in (A) and (B) indicate a significant improvement of \organa over the manual experiment for specified metrics\revise{}{, which was computed using a one-tailed T-test. If the p-value of the T-test was less than $0.05$, we noted the difference as significant.}
  }
  \label{fig:human-subjective-study}
\end{figure*}

\subsubsection*{User study -- quantitative results}

\cref{fig:pourbaix} compares Pourbaix plots generated by experimental chemists and \organa after the electrochemistry experiment. We compare parameter estimates based on combined data from all human experiments on one hand and all \organa experiments on the other. The values are comparable. For p$K_{\mathrm a1}$ \organa produces 8.03 and chemists 8.02. For the slope the estimated from \organa is -61.3~mV/pH ~unit, while for the chemists it is -62.7~mV/pH~unit.

    \cref{fig:interaction_time} depicts the time history and frequency of chemists' interactions with \organa across different experiment modes. Notably, \organa demonstrates superior performance in terms of human involvement. Manual experimentation \suggest{(\textbf{T1})}{} averaged over 30 minutes, while \organa, during the startup phase, required 7.35 minutes for written and 4.27 minutes for spoken instructions. \organa toubleshooting  took chemists an average of 1.30 minutes to provide feedback on errors. Additionally, the CLAIRify-style workflow \suggest{(\textbf{T4})}{} necessitated 17.65 minutes of user involvement on average.
In 3 experiments with correct results out of 40 total (8 users * 5 experiments without intentional bugs), \organa incorrectly alerted the human (false positive). Among 8 experiments with introduced errors, \organa failed to detect an issue in only one instance (false negative).

In \cref{fig:pourbaix}, the star symbol denotes an instance where \organa detected and alerted the user to an issue during a chemistry experiment, enabling timely correction. In contrast, the cross in the bottom left figure \suggest{(\textbf{T1})}{} underscores a scenario where a user omission in a manual user study became apparent only during subsequent data analysis.


\subsubsection*{User study -- qualitative analysis}

\cref{fig:human-subjective-study} presents qualitative survey results. In \cref{fig:human-subjective-study}(a), NASA-TLX responses for performing manual experiments \suggest{(\textbf{T1})}{} vs. \organa \suggest{(\textbf{T1-2})}{} reveal reduced demand and effort across all categories. 
Significantly, \organa halved participant frustration and reduced physical demand fourfold.
In \cref{fig:human-subjective-study}(b), we overlay results from the SUS for the manual experiment and \organa for each of the ten questions. We find that \organa shows significant improvement compared to manual in the following areas: desire to use the system frequently, reduced complexity of the system, improved consistency, and reduced cumbersomeness.
In \cref{fig:human-subjective-study}(c), responses to our custom questionnaire indicate unanimous agreement among chemists regarding the potential for automating repetitive lab tasks. Almost all chemists express comfort with robots performing some tasks, with diverse modalities for interaction. Nearly all find \organa easy to interact with (only one was neutral) and capable of accurately capturing intended experiments. Preferences for speaking vs. writing to \organa vary bimodally. While \organa is appreciated for reducing the time that humans need to be involved in the experiment, chemists prefer to be kept informed during experiment execution. The post-experiment summary is unanimously valued by all chemists.

\section*{DISCUSSION}
\label{sec:discussion}

\paragraph{Reliability and reproducibility of chemistry results}

\organa reliably reproduced results from literature for various experiments.
In the solubility experiment, \organa estimated compound solubility with $10.2 \pm 2.2\%$ mean and standard deviation values. In the electrochemistry experiment (\cref{fig:pourbaix}), \organa produced comparable results to \newsuggest{experienced}{} chemists, yielding a slope value of $-61.4 \pm 0.5~mV/pH~unit$ and a p$K_{\mathrm a1}$ value of $8.03 \pm 0.17$ across three runs of the experiment. These values are similar to values report in literature. While \organa demonstrated proficiency in reproducing literature results in various experiments, challenges persist in tackling complex chemistry tasks requiring advanced perception, manipulation, and planning. \suggest{}{Moreover, the use of specialized devices such as liquid and solid dispensers can improve the accuracy of the results whenever feasible.}

\paragraph{Modularity}
\organa is developed with modularity in mind, taking a step toward independent development and interconnection of modules to obtain and extend the overall functionality.
Instances of \organa (\cref{fig:fig4_old_experiments}) are applied in four chemistry experiments and integrate variations of natural language processing (NLP), perception, TAMP, robot execution, and data analysis, along with commonly available lab hardware. \revise{}{\organa can be used with a fundamentally different setup, for example, in experiments with continuous microfluidic platforms~\cite{liu2017microfluidics}.}
\suggest{This}{A} modular approach enables rapid customization for new applications, aligning with the concept of material on demand for lab automation.
Utilizing a general-purpose robot equipped with a multi-modal large perception and manipulation model can potentially enhance flexibility, reducing the need for specialized algorithms in novel applications~\cite{brohan2023rt}.
In electrochemistry experiments, we leverage LLMs and a transformer-based visual perception architecture for high-level task planning and object perception.
In contrast to other lab automation projects where robots operate in structured environments with fixed object poses, \organa relaxes this assumption. It perceives and acts in a semi-structured environment where objects and their poses can vary.

\paragraph{Evaluation of autonomy and robustness}

\organa successfully completed four different long-horizon chemistry experiments, including:
\begin{itemize}
    \item Solubility: 7 steps plan with 25.63 mins execution time performed for 2 times;
    \item Recrystallization:  8 steps plan with 44.80 mins execution time performed for 1 time;
    \item pH measurement:  6 steps plan with 3.85 mins execution time performed for 1 time;
    \item Electrochemistry: 114($6 \times 19$) steps plan with 130.00 mins execution time performed 2 times.
\end{itemize}

\looseness=-1
The human-in-the-loop feature of \organa makes it more robust to failures and reliable through timely interaction with users. In \cref{fig:pourbaix}, the star indicates a scenario where \organa detected and informed the user about an issue during a chemistry experiment, enabling timely correction.

\paragraph{Efficiency and support for \suggest{}{parallel experimentation}}
\organa increases efficiency and maximizes the usage of resources in chemistry experiments by parallelizing the execution of tasks using the available resources, i.e., pump, robot arm, stirrer, polisher, potentiostat, and pH-sensor. This is achieved by solving the TAMP and scheduling problems together, which can potentially lead to accelerated material discovery, a crucial component that involves supporting high-throughput experimentation and screening.
In this work, \tamp minimizes a cost function tied to total time; however, it is also possible to consider additional costs associated with the quality of task execution.
We demonstrated that \tamp results in a notable enhancement of 274 seconds (21.1\%) in the overall electrochemistry time compared to sequential task execution.
Solving TAMP and scheduling problems together incur an overhead compared to sequential TAMP during the planning phase. Employing learning-based techniques or LLMs is a potential approach for reducing planning time~\cite{khodeir2023learning, liu2023llm}.

\paragraph{Safety}
To address safety in lab automation, aside from relying on a human-in-the-loop for disambiguation, \organa relies on constrained motion planning, consistency checks, and feedback integration.

Constrained motion planning was applied and proved to be critical in solubility and recrystallization experiments in order to prevent spills while moving vials between locations. 
\organa, equipped with LLMs for reasoning, has the capability to detect unexpected events and address them through user interaction. This feature holds promise for enhancing safety when safety rationales are provided to it by the human.
Moreover, \organa's report generation feature can play a crucial role in documenting and keeping scientists informed in case of safety violations.
For example, in the report, a section could be dedicated to safety-related notes, according to safety rationales and metrics.
Finally, comprehensive safety should extend to both the physical and psychological well-being of humans, taking into account chemical, electrical, and mechanical hazards during synthesis and reactions. This necessitates preemptive and post-event safety measures in perception, planning, and execution, facilitating timely adjustments.

\paragraph{Interactions Between Chemists and \organa}
\looseness=-1
The user study indicates participants found lab automation, specifically \organa, useful. They expressed comfort with various communication modalities such as command line interface (CLI), GUI, or natural language. 

\looseness=-1
The study demonstrated a significant reduction in user physical load and frustration during the chemistry experiment with \organa. Participants consistently rated \organa as significantly useful (SUS questions 1-2) and expressed satisfaction (SUS questions 8-9) with its performance. The findings suggest that enhancing \organa's ease of use and learning could further streamline system usability, which also motivates the use of such systems for scientists with physical disabilities.

\looseness=-1
While users did not perceive a significant increase in efficiency based on workload and SUS studies, \cref{fig:interaction_time} indicates the quantitative importance of \organa in reducing human temporal workload. Specifically, for testing three buffer solutions, users saved 88.4 \% of time by interacting with \organa through audio compared to manual experimentation. The discrepancy between the human subjective and objective measures might stem from users performing only half of the full experiment manually (as opposed to the full experiment). Nevertheless, this result highlights the potential for system improvement.

Users agree on the necessity of keeping humans in the loop of autonomy. In fact, half the users expressed uncertainty in trusting a robot to complete the experiments autonomously from start to finish. This indicates the need for careful consideration when integrating robots into human workflows. Comprehensive report generation was identified as one potential method of increasing trust, as well as pinging humans while the experiment was being executed at moments of uncertainty. 

\looseness=-1
\paragraph{Limitations}
\organa currently relies primarily on independent sensor modalities for perception. There is potential in exploring multimodal perception to monitor the progress of chemistry tasks and enhance decision-making. 
An example can be found in~\cite{el2023keeping} to monitor chemistry task progress, where several process parameters (such as temperature and stir rate) and visual cues (such as volume, color, turbidity) were combined to control a chemistry process inside a reactor. Another instance of multimodal perception involves using haptic feedback and vision for object manipulation~\cite{murali2023touch, lee2019making}.
Finally, although we addressed the challenge of transparent object detection and pose estimation for our setup in electrochemistry experiments, there is much to be done to develop a robust model-free transparent object perception, considering their textureless and reflective surface~\cite{jiang2023robotic}.

A limitation of \tamp lies in the complexity of defining a PDDL domain for planning robot actions, potentially making it challenging for chemists who are not experts in planning and robotics to modify it.
Additionally, it lacks support for online replanning, limiting its adaptability to uncertainties in task execution. We are working on LLMs for task planning and replanning, which holds promise for solving multi-stage long-horizon tasks, but validating the proposed LLM plans and ensuring generalization remains a challenge~\cite{anonymous2023replan, liu2023llm, singh2023progprompt, liang2022code, driess2023palm}. In past work, we have started to address this by learning planning heuristics from experience ~\cite{khodeir2023learning, khodeir2023policy}.

\revise{}{
To fulfill the promise of flexible lab automation, the robotic agent should be able to automatically prepare the experimental setup, a capability currently lacking in the present work. For example, a mobile robot could retrieve clean beakers or vials from their shelves, insert pump tubes or sensory probes into the vials and beakers, and then proceed to run the experiment.}



\section*{EXPERIMENTAL PROCEDURES}
\label{sec:MaterialsAndMethods}

\subsection*{Resource availability}

\subsubsection*{Lead contact}

Requests for further information and resources should be directed to and will be fulfilled by the lead contact, Kourosh Darvish (kdarvish@cs.toronto.edu).

\looseness=-1
\subsubsection*{Materials availability}
This study did not generate new materials.

\subsubsection*{Data and code availability}

All the data required to evaluate the presented conclusions is available within the paper and in the Supplementary Materials. The dataset for perception evaluation can be found at \href{https://ac-rad.github.io/organa/}{https://ac-rad.github.io/organa/}. The code for \organa can also be found at \href{https://github.com/ac-rad/organa}{https://github.com/ac-rad/organa}.

\begin{figure*}[!htbp]
    \centering
    \includegraphics[width=0.8\textwidth]{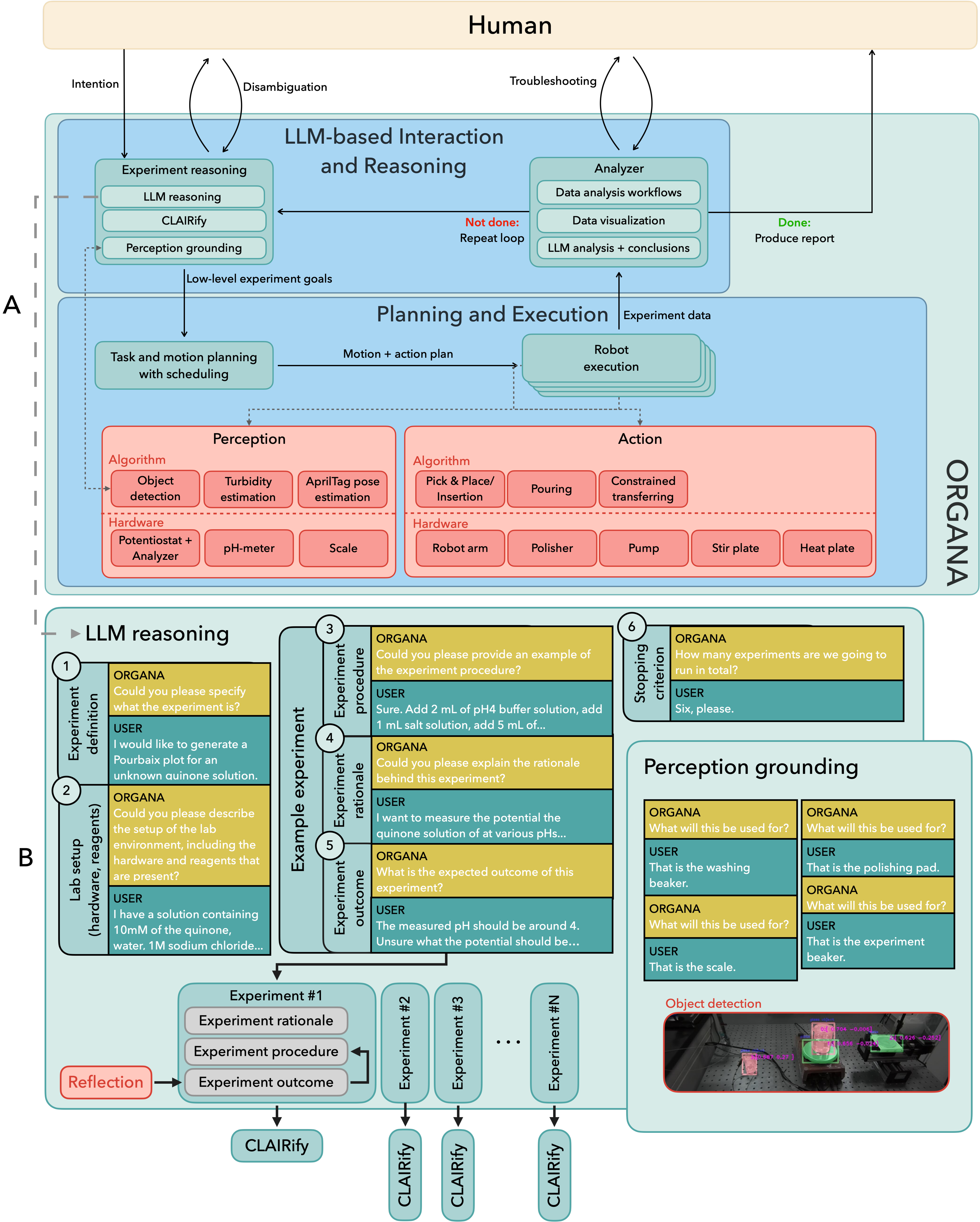}
    \caption{ \textbf{\organa's architecture and workflow.}
    \suggest{}{(A)} Users interact with \organa to convey their intentions for chemistry experiments and disambiguate the functionality of objects by grounding the scene.
    The LLM-based \nlp module translates these instructions into chemistry experiment plans and goals. Subsequently, \tamp generates parallel task and motion plans for execution, optimizing hardware utilization. \robot, equipped with action and perception skills, executes plans in parallel to maximize equipment usage and conduct experiments. The \analyzer processes raw data to estimate experiment progress and provides feedback to determine the next steps. Human notifications occur only when necessary to address unexpected situations. \suggest{}{(B) \nlp translates the experiment goal into a multi-step experimental plan by asking the user to define four key pieces of information: experiment definition (goal), lab setup (what hardware and reagents the robot can access), example experiment (how a user would explain the procedure for one experiment), and the stopping criterion. For each experiment, the procedure is processed by CLAIRify to generate TAMP input. Once an experiment is complete, the observations and outcomes are fed back into the planner to help plan for the next experiment (Reflection).  At the beginning of the experiment, the user is also asked to assign semantic meaning to the hardware that  \organa accesses during the experiment. This is so that \organa  knows what each vessel is used for if there are multiple of the same vessel type.}}
    \label{fig:fig2_architecture}
\end{figure*}

The architecture, workflow, and main components of \organa are described in \cref{fig:fig2_architecture}. The following sections elaborate on the details of each component.

\subsubsection*{Large Language Model (LLM)-Based Interaction and Reasoning}
\label{sec:method:nlp}



Generating low-level robot plans for each experiment can be \suggest{burdensome}{tedious} for chemists\suggest{}{, especially when several (potentially repetitive) experiments need to be conducted for the synthesis and characterization of materials.} 
To streamline this, \nlp facilitates the process through three steps:
i) \suggest{generating a natural language chemistry experiment task description from the intention of chemists for an experiment through an interactive conversation,}{generating a natural language chemistry experiment task description from the high-level instruction of chemists for multiples experiments through an interactive conversation at the start-up phase,}
ii) translating the experiment's natural description into valid structured language,
and iii) resolving ambiguities by interacting with the user during experiments, including grounding perception information and addressing unexpected outcomes.

\paragraph{Autonomous experiment reasoning}
\suggest{}{While previous works have shown that LLMs are able to generate plans for a single chemistry procedure, autonomously generating plans for a series of experiments is more challenging because the continuity and interdependencies between experiments require understanding complex sequences of actions and their consequences. This involves not only recalling completed experiments but also predicting outcomes and adapting strategies based on results from previous steps.  The goal of the start-up phase is to provide  \organa with enough information about experiment goals and the world state of the lab so it can autonomously perform a series of experiments instead of needing to provide \organa with step-by-step instructions.  During the start-up phase of the experiments, users interact with the system via text or speech. \suggest{The goal of this phase is for the system to identify a series of experiments that it will perform (see Figure 2b).} To propose a series of multi-step experiments, \nlp acquires information from the user about the following categories: experiment description, lab setup (available hardware, reagents), an example of how to execute one experiment (procedure, rationale, expected output), and stopping criterion. \suggest{}{An overview of this process is shown in \cref{fig:fig2_architecture}.}  The example experiment provided by the user is the basis for how \nlp constructs the remaining experimental procedures using CLAIRify. \nlp proposes subsequent experiment plans based on the experiment goal provided by the user and past experiments that have been completed.} \suggest{\nlp autonomously generates several experiments' natural descriptions by identifying user intentions. This includes inquiries user about experiment goals, expected observations, rationale, sample experiment procedures, and available reagents in the scene.Leveraging a summary of observations from past experiments and the current time step, \nlp employs}{} \suggest{}{This is accomplished by}  the ReAct prompting scheme~\cite{yao2023react}\suggest{}{, which generates} \suggest{to generate experimental plans using}{} \textit{(thought, action, observation)} tuples \suggest{}{after performing an experiment (the first tuple is parsed from the user input). In our case, an \textit{action} is viewed as an experimental plan, \textit{observations} are the measured experimental values post-plan execution, and \textit{thought} represents the rationale behind a given experiment.} \suggest{}{When the LLM is prompted to generate a new experiment, it has access to past \textit{(thought, action, observation)} tuples. This enables a chain-of-thought style reasoning, which provides the LLM context over what has been done already and why. This is important when planning multi-step experiments because past experiments need to be taken into account when proposing a new one to not repeat them unnecessarily or take into account any constraints.} \suggest{An \textit{action} is viewed as an experimental plan, \textit{observations} are the measured experimental values post-plan execution, and \textit{thought} represents the rationale behind a given experiment.}{} 

\suggest{}{After an experiment is complete, the results are analyzed. If the results do not match expected outcomes (as identified by the user in the experiment goal definition), the user is pinged to verify. \nlp incorporates this feedback, as well as outcomes from all past experiments, to plan subsequent ones.}

\suggest{}{In the startup phase, \organa also asks the user to semantically ground the hardware it perceives in the scene by identifying what each vessel is intended to be used for. \organa then labels each vessel with its name so that it can generate plans that utilize the correct vessel. This is important if there are multiple of the same vessel type in the experiment. \update{For example, if there are two beakers in the experiments (one used for the reaction and another used for waste), it is imperative that \organa does not use them interchangeably.}{In our electrochemistry experiment, one beaker is used for the reaction, and the other is used for waste. Therefore, it is imperative that \organa does not use them interchangeably.}} 
\update{}{However, if the beakers were interchangeable, \organa could use this information during planning to enable parallelization.}

\suggest{}{The \nlp-human interaction is shown in \cref{fig:fig2_architecture}(b).} Details about user interaction modalities \suggest{}{(text or speech)} and implementation specifics can be found in  \SINoteTWO and \SIFigFOUR.

\paragraph{Experiment description to valid structured task}
We employ CLAIRify~\cite{Skreta2023Errors} to convert the natural language description of a chemistry experiment into structured language codes in the XDL language, which are used as goals for planning by \tamp. CLAIRify utilizes an iterative prompting scheme to guarantee syntactic validity in the output language domain. \revise{}{Although there may be minor variations in the generated XDL scripts at each iteration of the experiment (such as different whitespace characters), the final plan for solving the PDDL problem remains the same.}

\paragraph{Human-in-the-loop~disambiguation~and~troubleshooting}
In addition to planning, \organa also engages with the user for scene clarification and resolving inconsistencies between \textit{expected} and \textit{current} observations\revise{w}{}. Following~\cite{majumdar2023findthis}, \organa addresses scene ambiguities during startup by grounding object functionalities. To handle unexpected experimental outcomes, a human-in-the-loop approach is adopted, where \nlp reasons over \textit{observations} and \textit{expected observations}, prompting the user to investigate and decide on further actions. Examples of ambiguity and uncertainty resolution and their prompts are detailed in \SINoteTHREE.

\subsubsection*{Task and Motion Planning with Scheduling}
\label{sec:method:tamp}

\begin{figure*}[t]
    \centering 

    \includegraphics[width=0.99\textwidth]{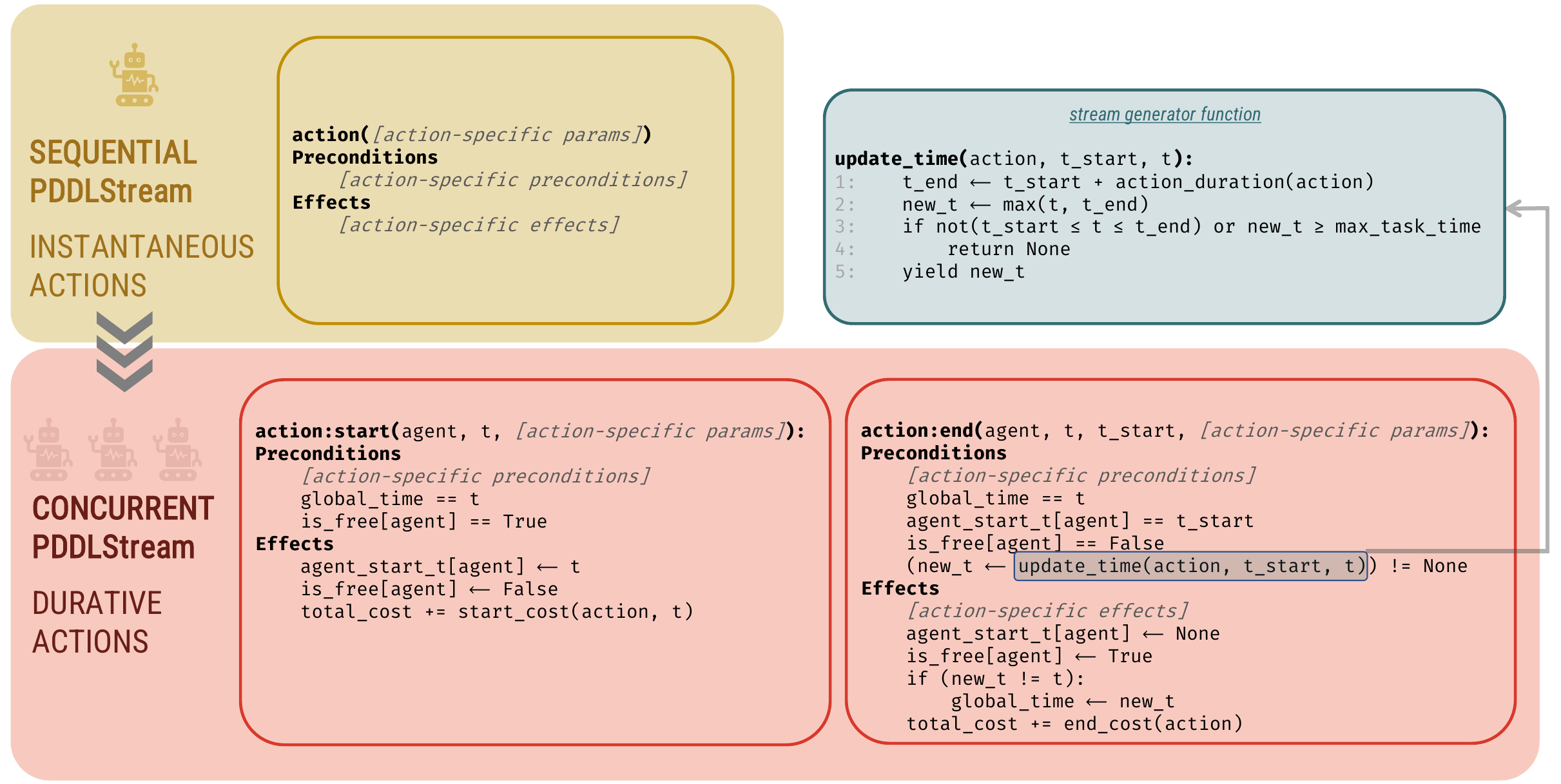}

    \caption{\textbf{Using PDDLStream to solve task and motion planning problems with scheduling.}
    To support parallel task execution in \organa, we transform the instantaneous actions $\texttt{action} \in \mathcal{A}$ (on top) to durative actions from PDDL2.1 with starting $\texttt{action:start}$ and ending $\texttt{action:end}$~\cite{fox2003pddl2} (at the bottom). For this purpose, the preconditions and effects are updated accordingly to meet the requirements and constraints of concurrent plans. For example, while the agent is acting, it cannot be assigned to any other actions ($\texttt{is\_free[agent]}$ is set to false as an effect of the starting action).
    In addition, non-negative $\texttt{global\_time}$ and $\texttt{agent\_start\_t[agent]}$ literals are added to keep track of the current global execution time and the starting time of each action by the agent. Two stream functions $\texttt{start\_cost(action, t)}$ and $\texttt{end\_cost(action)}$ are added to provide the cost of action starting and ending.
    Moreover, $\texttt{update\_time()}$ function, in the precondition of $\texttt{action:end}$, is associated with a stream generator that gets as input the action, its starting time, and the current global execution time and it updates the current global execution time.
   The constraints on $\texttt{update\_time()}$ stream are added in line 3 of the stream generator function on top right, ensuring the current global execution time is higher than the action starting time, and the action is ended with the correct duration. Moreover, it ensures the updated time is less than the maximum time allowed to reach the goal of PDDLStream. \SINoteEIGHT details these changes using PDDL syntax.}
    \label{fig:TemporalPDDLStreeam-Actions}
\end{figure*}

To \suggest{enable high-throughput experimentation}{speed up experimentation}, the TAMP planner should facilitate parallel task execution by robots and other resources or equipment.
We adapted PDDLStream~\cite{Garrett2020PDDLStream} with PDDL2.1~\cite{fox2003pddl2} to support durative actions and introduced a time-variant cost function to enhance task execution efficiency. 
PDDLStream is represented by the tuple $<\mathcal{P}, \mathcal{A}, \mathcal{I}, \mathcal{G}, \mathcal{S}>$, respectively defining predicates, actions, initial state, goal state, and streams. The stream $S(\textbf{x})$ over a tuple of literals $\textbf{x}$ acts as a conditional sampler, declaring the satisfaction of the relation between its input and output tuples.
To enable durative actions, $\texttt{a} \in \mathcal{A}$ is substituted with \textit{starting} and \textit{ending} actions, denoted as \revise{$\texttt{a-start}$}{$\texttt{a:start}$} and \revise{$\texttt{a-end}$}{$\texttt{a:end}$}~\cite{fox2003pddl2}.
Additionally, the starting time of the $i$'th action in the plan $\pi$, where $a_i \in \pi$, is linked to \revise{$t_{\texttt{a-start},i}$}{$t_{\texttt{a:start},i}$}, and its duration is indicated by $D_{\texttt{a},i}$.
For efficiency and reduced task execution time, the total cost is defined as:
\begin{equation}
    \texttt{total\_cost} = \sum_{i=1}^{k} (t_{\texttt{a:start},i} + D_{\texttt{a},i})~~~ \forall \texttt{a} \in \pi, k=|\pi| .
\end{equation}
\cref{fig:TemporalPDDLStreeam-Actions} details the transformation of action $\texttt{a}$ to \revise{$\texttt{a-start}$}{$\texttt{a:start}$} and \revise{$\texttt{a-start}$}{$\texttt{a:end}$}, along with updates to their cost functions.
\SINoteEIGHT provides an overview of PDDL syntax and details the modifications to action descriptions in the PDDL language.


\begin{algorithm}[H]
\scriptsize
\begin{algorithmic}[1]
\caption{\textsc{Temporal-PDDLStream}}
\label{alg:temporal-PDDLStream}
\Require {$\mathcal{A}, \mathcal{S}^o, \mathcal{S}^c, \mathcal{I}, \mathcal{G} $ }
\Ensure {$\pi$}

\State $ \mathcal{U}^c = $ \textsc{ApplyStreams($\mathcal{S}^c, \mathcal{I}, next$)}  \Comment{eagerly evaluate costs}
\While{$True$} 

\State $ \mathcal{U}^* = $ \textsc{ApplyStreams($\mathcal{S}^o, \{\mathcal{I}, \mathcal{U}^c\},$OptOutput)}  \Comment{optimistic stream}
\State $ \pi^* = $ \textsc{Search($\mathcal{A}, \{\mathcal{I}, \mathcal{U}^c, \mathcal{U}^*\},\mathcal{G}$)}  
\State $ \pi, \psi = $ \textsc{Evaluate($\{\mathcal{I}, \mathcal{U}^c\}, {\mathcal{U}^*},\pi^*, \mathcal{G}$)}
\IfThen{$\pi \neq$ None}{\Return} $\pi$
\EndWhile

\end{algorithmic}
\end{algorithm}

In \citet{Garrett2020PDDLStream}, various methods to solve the PDDLStream problem are discussed, including an incremental approach where streams are certified eagerly and blindly before the search, leading to inefficiency due to the generation of irrelevant facts during stream evaluation.
Another method involves optimistic certification of streams, enabling a lazy exploration of candidate plans. While more efficient, this approach does not support time-varying functions with streams.
In Temporal TAMP, \suggest{}{time-variable} costs \suggest{, varying with time,}{} are updated through streams. To overcome the limitations of the two existing methods in solving the temporal TAMP problem, we integrate both approaches. Specifically, time-varying streams linked to cost functions and timings are evaluated eagerly, while the remaining streams are evaluated optimistically.
Additionally, by imposing reasonable constraints on eagerly evaluated streams, we restrict the search space to enhance search efficiency.
\cref{alg:temporal-PDDLStream} outlines our method for solving temporal PDDLStream problems, taking input durative actions $\mathcal{A}$, optimistic streams $\mathcal{S}^o$, cost-related eager streams $\mathcal{S}^c$, and initial states $\mathcal{I}$ with a goal $\mathcal{G}$.
Initially, we assess time- and cost-associated streams, incorporating them into the current set of certified facts $\mathcal{U}^c$.
The remaining streams $\mathcal{S}^o$ are optimistically evaluated using the \texttt{OptOutput} procedure to create an optimistic object tuple $u^* \in \mathcal{U}^*$.
In line 4, we employ a fast downward planning system utilizing weighted A* heuristic search~\cite{helmert2006fast} to find an optimistic plan $\pi^*$ with a focus on minimizing the plan cost.
Finally, the optimistic plan and its associated streams are evaluated and certified, returning upon finding a plan. \revise{}{\organa relies on the TAMP planner described here to manage all agent plans. The preconditions for each step of the experiment are defined to ensure that no action is executed until its preconditions are satisfied. The scheduler returns a feasible solution that accounts for all conditions and prevents race conditions.}

\subsubsection*{Perception}
\label{sec:method:perception}

To achieve autonomous chemistry experiments, we suggest a dual-level perception framework. The first level monitors chemical task progress by characterizing materials, while the second level focuses on perceiving workspace objects for robot manipulation. Our approach integrates various sensors and utilizes perception algorithms for monitoring reactions and estimating workspace states.
Details of perception algorithms are provided below, and hardware specifics can be found in  \SINoteFIVE.

\paragraph{Turbidity visual feedback}
Turbidity, indicating solution opaqueness, gauges undissolved solvent in solubility experiments. Following HeinSight \cite{zepel_lai_yunker_hein_2020}, we adopt the solution's average brightness, observed by a robot with an in-hand camera, as a proxy for turbidity. Using the Hough Circle Transform~\cite{illingworth1988survey}, the robot identifies the dish in a top-down view, extracts the minimum square region enclosing it in hue, saturation, and 
value (HSV) color space, and computes the average brightness as the turbidity value. See \cref{fig:fig4_old_experiments} for an automated turbidity measurement example.

\paragraph{Transparent and opaque object detection and pose estimation}
Perceiving transparent objects poses challenges due to violations of the Lambertian assumption and the textureless nature of transparent surfaces\cite{jiang2023robotic}. To address this, we integrated cutting-edge perception algorithms in \perception for effective transparent object detection and pose estimation in the scene (refer to \SIFigTWO for a visualization of the pipeline).
First, we employed Grounding DINO~\cite{liu2023grounding}, a transformer-based zero-shot object detection model ~\cite{zhang2022dino} with grounded pre-training. This model enables flexible object detection by accepting human category names and images as input and providing bounding box information, labels, and confidence for each detection.
Next, we applied Non-Maximum Suppression (NMS), a standard post-processing method in object detection \cite{10028728}, to remove duplicate detections and retain the most pertinent bounding boxes.
The resulting bounding box was then fed into the Segment Anything model (SAM) \cite{kirillov2023segment}, utilizing a transformer-based image encoder and mask decoder for segmentation. SAM offers options for points, boxes, and segment prompting through a CLIP-based prompt encoder~\cite{radford2021learning}. Further implementation details can be found in 
  \SINoteONE, \SIFigPERCEPTION, and \SITableONE.

Estimating 3D object poses from 2D segments requires depth information for reprojection. We obtained the necessary depth data using the ZED camera depth map \cite{Stereolabs}.
Opaque object point clouds maintain high accuracy, but transparency introduces distortion, capturing background surfaces. To mitigate this, a practical solution involves estimating the camera distance to the front surface of transparent objects by filtering the least distorted closest 10\% of points.
The minimal 3D bounding box was derived from each object's point cloud convex hull using Open3D functions \cite{Zhou2018}. Principal Component Analysis (PCA) \cite{labrin2020principal} identified the object's major axes, forming the rotation matrix of the object's frame of reference. Finally, object pose in the robot world frame was determined through extrinsic camera calibration.

\subsubsection*{Skills for Chemistry Experiment}
\label{sec:skills}

To enable autonomous chemistry experiments, it is crucial to integrate a diverse set of robot skills and laboratory tools. Skills are attained through either specialized hardware (refer to  \SINoteSIX) or diverse algorithms, as detailed below.

\paragraph{Pick \& place and insertion skills}
\organa uses object pose information coming from perception to determine the robot end-effector's target frame. These frames are used for grasping, placement, and insertion. To enhance robustness and avoid collisions, a pre/post pose strategy is applied, such as pre-insertion, insertion, and post-insertion poses during object manipulation. The robot joint trajectory is computed using inverse kinematics~\cite{beeson2015trac} and the probabilistic roadmap (PRM*) path planning~\cite{Kingston2019Exploring}.

\paragraph{Constrained motion planning skill}
In a chemistry lab, a common task involves transporting containers with liquids and powders. In experiments, especially those related to solubility and recrystallization, we introduced orientation constraints to prevent spillage when the robot transfers beakers. We utilized the PRM* sampling-based method for robot motion planning~\cite{karaman2011sampling}. To incorporate $k$-dimensional path constraints $\mathcal{F}(q): \mathcal{Q} \to \mathbb{R}^k$ in the configuration space $\mathcal{Q}$, we applied a projection-based method to identify configurations satisfying constraints during PRM* sampling~\cite{Kingston2019Exploring}. When sampling a free configuration in PRM*, its projected value in the constrained configuration space is determined by iteratively minimizing $\mathcal{F}(q)$ using its Jacobian. 
For details on our implementation and evaluation, see~\citet{AuRo2023Clairify}.

\paragraph{Liquid and granular material pouring skill}
In the chemistry lab, liquid and granular solid pouring is routine, with liquid transfer handled by a pump. Precise powder pouring is challenging and expensive with existing hardware solutions. Inspired by manual pouring skills, \organa implements a robotic pouring technique in solubility and recrystallization experiments. This skill incorporates weight feedback using a proportional derivative (PD) controller and a shaping function, taking the desired substance target value as input and providing the desired rotational velocity of the robot end-effector. Additional details are available in~\citet{AuRo2023Clairify}.

\suggest{}{\paragraph{Electrode polishing skill}
In our electrochemistry experiment, \organa utilizes a polishing station to refine the glassy carbon electrode. This station comprises a polishing pad connected to two linear actuators that execute planar motion. A mechanical impedance, facilitated by a spring linked to the electrode jig, governs the normal interaction force between the polishing station and the robot end-effector. The electrode is polished with a circular motion realized by sending signals to the linear actuators. Further details on the polishing process are available in~\cite{yoshikawa2023polish}.}\newsuggest{}{This design is inexpensive and enables polishing tasks to be performed by a simpler robotic platform in the future, allowing the more dexterous robot to carry out other tasks.}

\subsubsection*{Automated Data Analysis and Report Generation}
\label{sec:analyzer-report-generation}
To enable \organa to generate comprehensive user reports, we integrated the following tools into \analyzer.

\subsubsection*{Electrochemistry Parameter Estimation}
\label{sec:parameter-estimation}
In the electrochemistry experiment, we aim to characterize the relationship between the pH and the redox potential, which is the potential that drives the reduction or oxidation half-reaction of a compound measured against a standard reference half-cell~\cite{goldbook_redox}.
We know that the relationship has three distinct regions of linear dependency, demarcated by pH values p$K_{\mathrm a1}$ and p$K_{\mathrm a2}$. We also know that the slope of the second region (from p$K_{\mathrm a1}$ to p$K_{\mathrm a2}$) is one-half of the slope of the first one (before p$K_{\mathrm a1}$), while in the last region (after p$K_{\mathrm a2}$) the redox potential does not change (slope is equal to zero). The model is therefore fully defined with 4 parameters: two inflection points (p$K_{\mathrm a1}$ and p$K_{\mathrm a2}$), a single slope variable (in our case $k$, the slope in region [p$K_{\mathrm a1}$, p$K_{\mathrm a2}$]) and one variable to define the redox potential offset (we take $\text{E}_{\text{inf}}$, the value in the third static region).

To produce parameter values based on collected data we utilize maximum likelihood estimation (MLE). In addition to that, we aim to produce an updated belief about the parameter values and the model line after each new data point has been sampled. We utilize this posterior over parameters to plot marginal distributions for each individual one  (see \cref{fig:electro} and \cref{fig:pourbaix} for examples). With an automated system, 
this is an important element in keeping the chemist aware of the current progress of the experiment, in order to catch any issues that the system might not automatically detect and in general to make any corrections necessary. Additionally, while utilizing an optimization algorithm to choose points to sample was not needed in the specific electrochemistry experimental setup, this output would allow for the simple application of out of the box optimization algorithms in the future. We give details of the method for estimating the posterior distribution in  \SINoteSEVEN.

\subsubsection*{Electrochemistry Report Generation}
\label{sec:report_gen}
\organa automatically generates a PDF summary report at the experiment's conclusion, offering a comprehensive overview of the chemistry experiment, including automatically-generated statistical analyses of the measurements, to the users. The report includes experiment details, logs of failures with corresponding resolutions, and summary plots analyzing the results. An example electrochemistry report is in 
 \SINoteNINE and \SIFigREPORTS.

\section*{Supplemental information index}

Supplemental PDF contains notes on experimental details, corresponding figures, and a table. \SIVideoONE is a live demonstration of \organa interacting with human and executing electrochemistry experiment in parallel.

\section*{Acknowledgments}
We thank members of the Matter Lab for participating in the user study. We would like to also thank Jinbang Huang, Lasse Bjørn Kristensen, and Jason Hattrick-Simpers for their insightful discussions. This research was undertaken thanks in part to funding provided to the University of Toronto's Acceleration Consortium from the Canada First Research Excellence Fund, grant number CFREF-2022-00042. We acknowledge the generous support of Dr. Anders G. Frøseth, the Acceleration Consortium, the Vector Institute, Natural Resources Canada and the Canada 150 Research Chairs program. F.S. and A.G. would also like to acknowledge the Discovery Grant.

\section*{Author contributions}
K.D. led the technical development and validation of the system, as well as TAMP with the scheduling.
M.S. created the human-\organa interaction and reasoning pipeline, led the user study, and created the report generation capability.
Y.Z. led system integration, developed the perception pipeline, and contributed to TAMP with scheduling.
N.Y. developed the chemistry experiment protocol and contributed to system development.
S.S. contributed to developing perception and system integration.
M.B. contributed to developing parameter estimation.
H.H. and Y.C. contributed to the chemistry experiment protocol.
H.X. contributed to developing perception.
A.A.G. supervised chemistry experiment protocols and speech interaction.
A.G. supervised large language model-based planning and reasoning.
F.S. supervised task and motion planning with scheduling, user study, and automated report generation. 
K.D., M.S., Y.Z., N.Y., S.S., and M.B. wrote the initial draft.
All the authors proofread the manuscript.

\section*{Declaration of interests}
The authors declare no competing interests.

\bibliography{references}

    \newpage
    \onecolumn
    \section*{\large\textbf{Supplemental information}}
\renewcommand{\thefigure}{S\arabic{figure}}  
\renewcommand{\thetable}{S\arabic{table}}    
\setcounter{figure}{0}  
\setcounter{table}{0}   

\newcounter{note}
\renewcommand{\thenote}{S\arabic{note}}
\newcommand{\notelabel}[2]{%
    \refstepcounter{note}
    \label{note:#1}
    {\large\textbf{Note \thenote: #2}}
}

\newcounter{video}
\renewcommand{\thevideo}{S\arabic{video}}
\newcommand{\videolabel}[2]{%
    \refstepcounter{video}
    \label{video:#1}
    {\large\textbf{Video \thevideo: #2}}
}

\notelabel{supplementary_results:perception-analysis}{Perception Analysis}
\addcontentsline{toc}{subsection}{\cref{note:supplementary_results:perception-analysis}}

\paragraph{Perception for Lab Automation Background}
Effective scene perception is paramount in the lab automation context, with a particular emphasis on two facets of perceptual skills: chemistry-level perception for synthesis monitoring and analysis and object-level perception for the manipulation of lab equipment. Addressing chemistry-level perception, early efforts, exemplified by LabPics \cite{LabPic}, introduced an innovative image dataset and employed convolutional neural networks (CNNs) to discern material phases and delineate phase boundaries. Recent advancements by HeinSight systems \cite{D3SC05491H, zepel_lai_yunker_hein_2020} have furthered this field by presenting a more generalized vision system that not only classifies material phases but also quantifies physical properties such as volume, color, and turbidity, thereby facilitating comprehensive monitoring and control of experimental processes. Moreover, in the work of \citet{D3DD00109A}, liquid viscosity estimation was achieved through the utilization of robot manipulators for collecting fluid motion videos, followed by 3D CNN analysis.

\looseness=-1
Turning to object-level perception, the transparency of many chemistry lab tools poses a unique         challenge. Notably, works have focused on detecting transparent vessels in occluded scenes by leveraging 3D reconstruction and multiview perception~\cite{xu2021seeing, wang2023mvtrans}. This approach enabled the estimation of depth, segments, and object poses, enhancing the efficacy of downstream manipulation tasks. These models focus on detecting common transparent and opaque objects; however, they do not generalize well to unfamiliar and novel objects. We adopted Grounding DINO~\cite{liu2023grounding} with SAM~\cite{kirillov2023segment}, a more versatile approach that is also capable of accepting unconstrained text prompts.  

\paragraph{Evaluation of object detection and pose estimation in \organa}
The perception pipeline is evaluated in two aspects: object detection and object position estimation. Average Precision (AP) was used to measure the detection performance for each object class. Mean Absolute Error (MAE) was used to evaluate the position estimate of the objects. Since we assumed that objects maintain an upright orientation and object orientation is not utilized in our chemistry experiments, the quality of the orientation prediction was not evaluated.  A real dataset related to the electrochemistry setup was constructed for conducting these evaluations.  

\begin{figure}[h!]
    \centering
    \includegraphics[width=1\columnwidth]{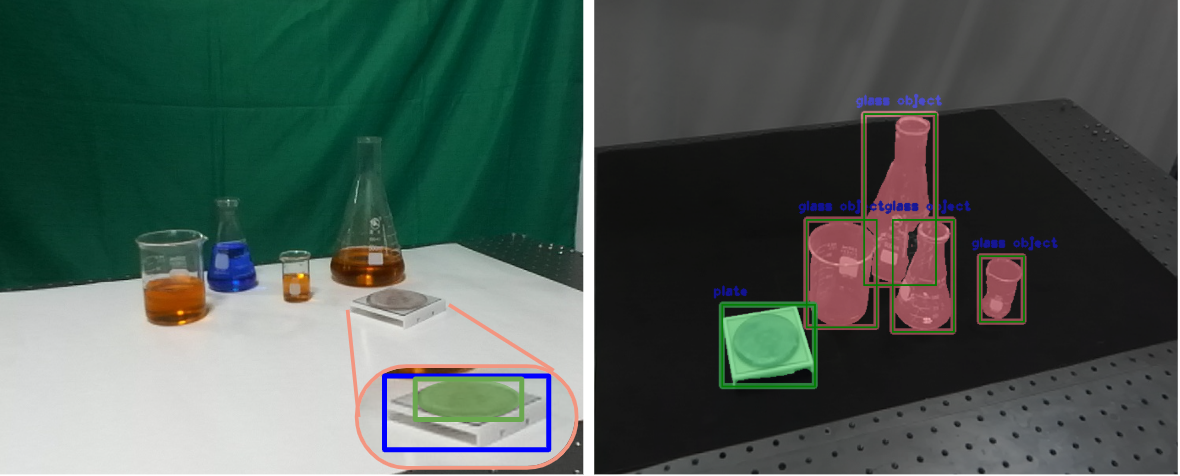}
    \caption{\textbf{Evaluation of \perception on two sample images.} Left: A sample image showing a possible object arrangement. For the polishing pad, the blue bounding box represents the ground truth and the green box is the predicted output from the perception pipeline. Right: An example of detection and segmentation of partially occluded vessels and the polishing plate.}
    \label{fig:sample_data_image}
    \addcontentsline{toc}{figure}{\cref{fig:sample_data_image}}
\end{figure}

\paragraph{Data Collection} 
We used a ZED Mini camera~\cite{zedmini} to collect images of varying scene setups. Specifically, chemistry equipment, including beakers, flasks, and a polishing pad were placed on a lab bench; we randomized lighting conditions, table backgrounds, colored liquids in transparent vessels, and object locations. We automated the data collection by having the robot with an eye-in-hand camera capture RGBD images from different angles, as shown in \cref{fig:sample_data_image}. In total, the dataset consists of 135 RGBD images captured across 17 different scenes. Each scene contains 4 transparent objects and 1 polishing plate.  

To obtain the ground truth object pose in the world frame, AprilTags \cite{olson2011apriltag} were randomly placed on the table. An image was then captured by the robot camera and the 6D poses of the tags were estimated using OpenCV functions \cite{opencv_library}. We converted these poses into world coordinates based on known camera poses and subsequently replaced the tags with associated objects on the table. For obtaining the ground truth 2D bounding boxes for each object, we created 3D mesh models based on measured dimensions. Point clouds of these mesh models were projected to 2D image space using the corresponding tag pose, camera intrinsic and extrinsic matrices. The 2D bounding box for each set of projected points was generated and treated as the ground truth.

\begin{figure*}[t]
    \centering
    \includegraphics[width=1.0\textwidth]{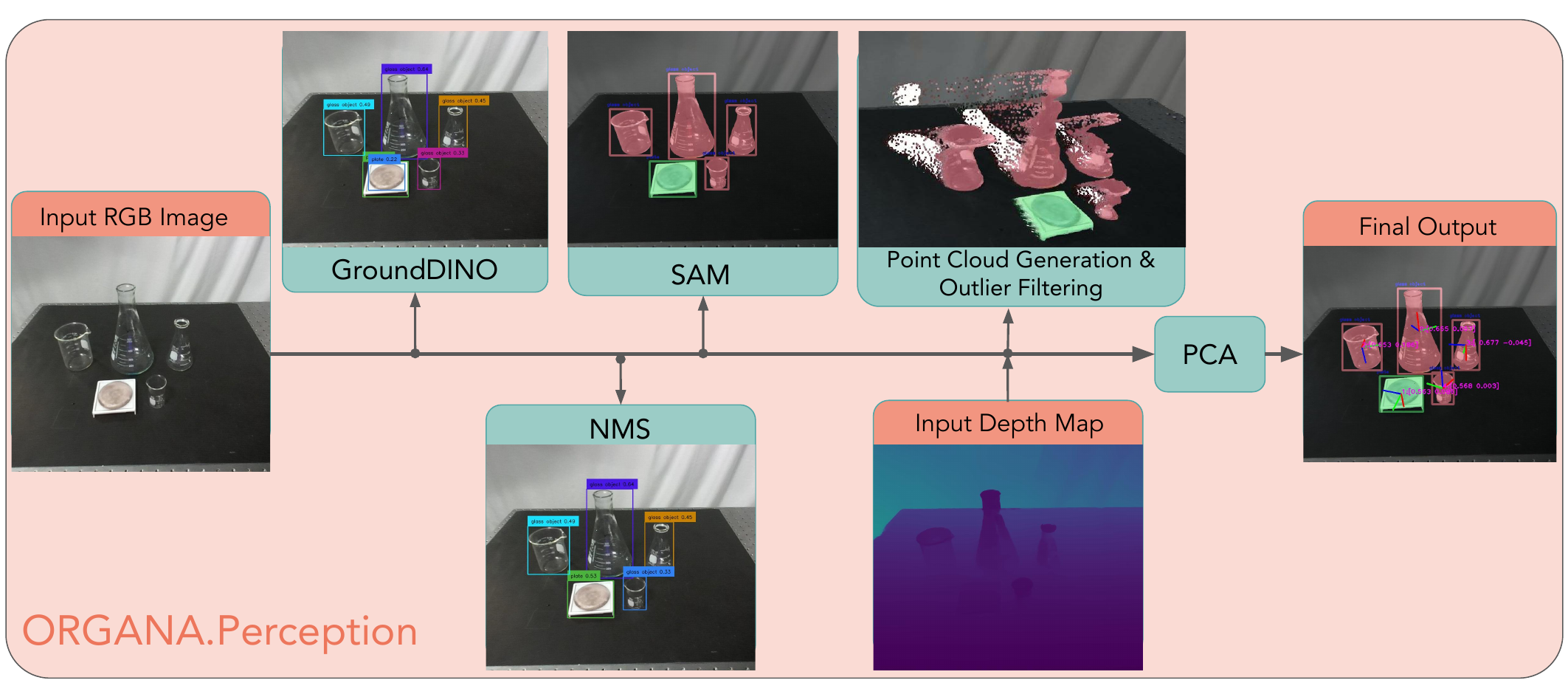}
    \caption{\textbf{\perception Pipeline.} Given an input RGB image, Grounding DINO detects objects of interest based on pre-defined prompts and predicts their 2D bounding boxes. NMS is then applied to remove redundant predictions, followed by SAM, which generates a segmentation mask for each detected object. To estimate object pose, the input depth map is used to generate 3D point clouds for each object, which are then filtered to remove outlier points. The object position is estimated based on the smallest 3D bounding box fitted to each set of point clouds, and the object rotation is estimated using PCA.}
    \label{fig:perception_pipeline}
    \addcontentsline{toc}{figure}{\cref{fig:perception_pipeline}}
\end{figure*}

\paragraph{Implementation} 
The perception pipeline was built based on Grounding DINO\cite{liu2023grounding} and SAM models\cite{kirillov2023segment} with a custom object pose estimation mechanism shown in \cref{fig:perception_pipeline}. For evaluation, we directly used the checkpoint of the Grounding DINO model  with the Swin-T backbone \cite{hwang2022tutel} and SAM with the ViT-B backbone \cite{50650}. To detect beakers and the polishing station in the scene, we prompted the Grounding DINO with ``glass object'' and ``plate''.

\paragraph{Results} 
For object detection,  \cref{tab:AP_table} summarizes the Average Precision (AP) for each class supported by the perception pipeline. We observed that the precision of glass objects is consistently high across different Intersection over Union (IoU) thresholds. In contrast, the pipeline achieves a high AP for detecting plates with low IoU thresholds, but the detection performance dramatically decreases as IoU increases. This occurs because the Groundning DINO model often recognizes the brown pad as the plate rather than the entire polishing pad fixture, resulting in a lower IoU, as demonstrated in \cref{fig:sample_data_image}. Additionally, we found that the pipeline can handle partial occlusion, as shown in \cref{fig:sample_data_image}. This capability enables the pipeline to withstand more realistic and complex conditions outside of preset object configurations.

\begin{figure}[h!]
    \centering
    \includegraphics[width=1\columnwidth]{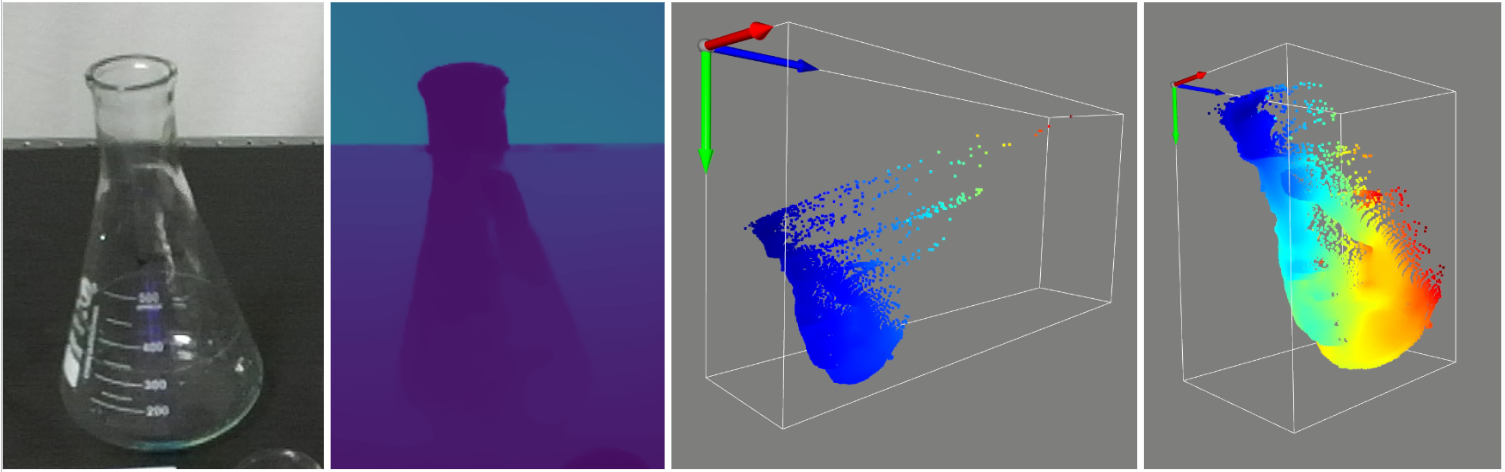}
    \caption{\textbf{Visualizations of the RGBD image and 3D point clouds for an Erlenmeyer flask.} The two figures on the right display the point cloud generated from the raw sensor depth map and the point clouds after outlier removal. }
    \label{fig:point_cloud}
    \addcontentsline{toc}{figure}{\cref{fig:point_cloud}}
\end{figure}

In terms of object position estimation, the pipeline achieved a Mean Absolute Error (MAE) of 3.5 cm, averaged over all objects in the dataset. Specifically, the small beaker has the lowest MAE of 2.4 cm, while the large flask has the highest MAE of 5.1 cm. The accuracy of the object position is also highly dependent on the depth map from the ZED camera, as discussed in the paper (``Transparent and opaque object detection and pose estimation" section). Although the ZED Mini camera is reported to have a depth error of 1.5\% within its range of 10 cm to 3 m \cite{zedmini}, the depth accuracy of transparent objects degrades given that background depth values are reported instead. To address this issue, we applied the radius outlier removal method \cite{Zhou2018} directly on point clouds as demonstrated in \cref{fig:point_cloud}, improving the overall MAE from 4.5 cm to 3.5 cm. Future work will involve developing the module to support a diverse set of lab equipment while achieving high precision in pose estimation with image input.

\begin{table}[h!]
    \centering
    \caption{Average Precision (AP) of the perception pipeline on each of the supported object classes for different IoU thresholds}
    \begin{tabular}{|c|c|c|}
    \hline
    IoU Threshold & AP of glass & AP of plate \\ \hline
    0.25          & 94.9        & 81.4        \\ \hline
    0.50          & 93.9        & 38.6        \\ \hline
    0.75          & 90.7        & 37.1        \\ \hline
    \end{tabular}
    \label{tab:AP_table} 
    \addcontentsline{toc}{table}{\cref{tab:AP_table}}
\end{table}

\newpage
\notelabel{appendix:nlp}{\nlp Implementation and Prompts}
\addcontentsline{toc}{subsection}{\cref{note:appendix:nlp}}
\paragraph{Text and speech interface}
Users can interact with \nlp using a text interface or speech. The speech interface detects when humans are speaking using the SpeechRecognition\footnote{\url{https://github.com/Uberi/speech_recognition}} library in Python. Speech is processed to text using OpenAI's Whisper model\footnote{\url{https://github.com/openai/whisper}}. OpenAI's GPT-4 model\footnote{\url{https://openai.com/gpt-4}} is used to reason over the user input and generate follow-up questions, which are processed to speech using ElevenLabs voice generators\footnote{\url{https://elevenlabs.io}}.

\paragraph{\nlp prompts}
The chemist interacts with \organa in three key stages: [1] Experiment Start-Up, [2] Physical Grounding, [3] Experiment. These stages are described below along with the LLM prompts. 

\paragraph{Phase 1: Experiment Start-Up}
The goal of \nlp during the Experiment Start-Up phase is to gather enough information in order to be able to execute the experiment. \nlp has a series of questions that need to be answered in order to perform an experiment. \nlp asks these questions to a user and deduces the answers based on their responses according to the template below:

\begin{lstlisting}[language=Python]
start_up_prompt = """
You are a robot chemist. A chemist will ask you to perform an experiment. Your goal is to find out all the information you need about the experiment. You fill out the following items:::

[Item 1] What the experiment is:::
[Item 2] The setup of the lab environment (what hardware and reagents are present):::
[3] An example of how to run the experiment
[Item 3a] The rationale behind the example experiment:::
[Item 3b] How you would do the example experiment:::
[Item 3c] What the expected outcome of the example experiment is:::
[Item 4] Number of experiments to be run in total:::

You will receive input from the user. You NEED to return two things:
[Item X] where X is the number of the item satisfied or [Item None] if no items were satisfied
[Question] The question you want to ask next. You must only ask one question at a time. If all the items have information, write <DONE>.
"""
\end{lstlisting}

\nlp prompts the chemist until all questions are satisfied. Below is an example of a completed prompt, collected from a user speaking:

\begin{lstlisting}[language=Python]
start_up_prompt = """
You are a robot chemist. A chemist will ask you to perform an experiment. Your goal is to find out all the information you need about the experiment. You fill out the following items:::

[Item 1] What the experiment is:::I would like to run a chemistry experiment. I would like to generate a Pourbaix plot for an unknown quinone
[Item 2] The setup of the lab environment (what hardware and reagents are present):::I have a solution containing 10mM of the quinone, water, 1M sodium chloride solution, and a series of 0.5 M pH X buffer solution (X is an integer between 4 to 9)two beakers (experiment beaker, waste beaker), a pH probe and potentiostat
[3] An example of how to run the experiment
[Item 3a] The rationale behind the example experiment:::I want to measure the potential of the quinone solution at various pHs. I will start at pH 4 and take a measurement there. The experiments should measure the potential at all pHs from 4 to 9.
[Item 3b] How you would do the example experiment:::Add 2 mL of pH 4 buffer solution. Add 1 ml NaCl solution. add 5 mL of water followed by 2 mL of quinone solution. Then measure the pH and run a CV scan. transfer r contents of beaker to waste beaker
[Item 3c] What the expected outcome of the example experiment is:::The measured pH should be around 4 (a little difference is okay). Unsure what the potential should be since it's the first experiment, but the trend is that potential should become more negative as pH goes up.
[Item 4] Number of experiments to be run in total:::6

You will receive input from the user. You NEED to return two things:
[Item X] where X is the number of the item satisfied or [Item None] if no items were satisfied
[Question] The question you want to ask next. You must only ask one question at a time. If all the items have information, write <DONE>.
"""
\end{lstlisting}

The answers are then processed into a dictionary of initial conditions:

\begin{lstlisting}[language=Python]
init_conditions_dict = {"goal": ... #[Item 1] from start_up_prompt,
                        "setup": ... #[Item 2] from start_up_prompt,
                        "thought": ... #[Item 3a] from start_up_prompt,
                        "action": ... #[Item 3b] from start_up_prompt,
                        "expected_obs": ...#[Item 3c] from start_up_prompt,
                        "num_repeats": ... #[Item 4] from start_up_prompt
                        }
\end{lstlisting}

Based on the answers to the prompts, \nlp then selects the experiment category. This is necessary to determining post-processing functions for experiment results. 

\begin{lstlisting}[language=Python]
experiment_type = """
Based on this experiment goal: {0}, what type of experiment is being done? Select type from [POURBAIX, TITRATION, SOLUBILITY, RECRYSTALLIZATION, NONE]. Only return type, nothing else. If you are uncertain, return NONE
"""
\end{lstlisting}

If the experiment type is not known, \organa exits. 

\begin{figure}[h!]
    \centering
    \includegraphics[width=1\columnwidth]{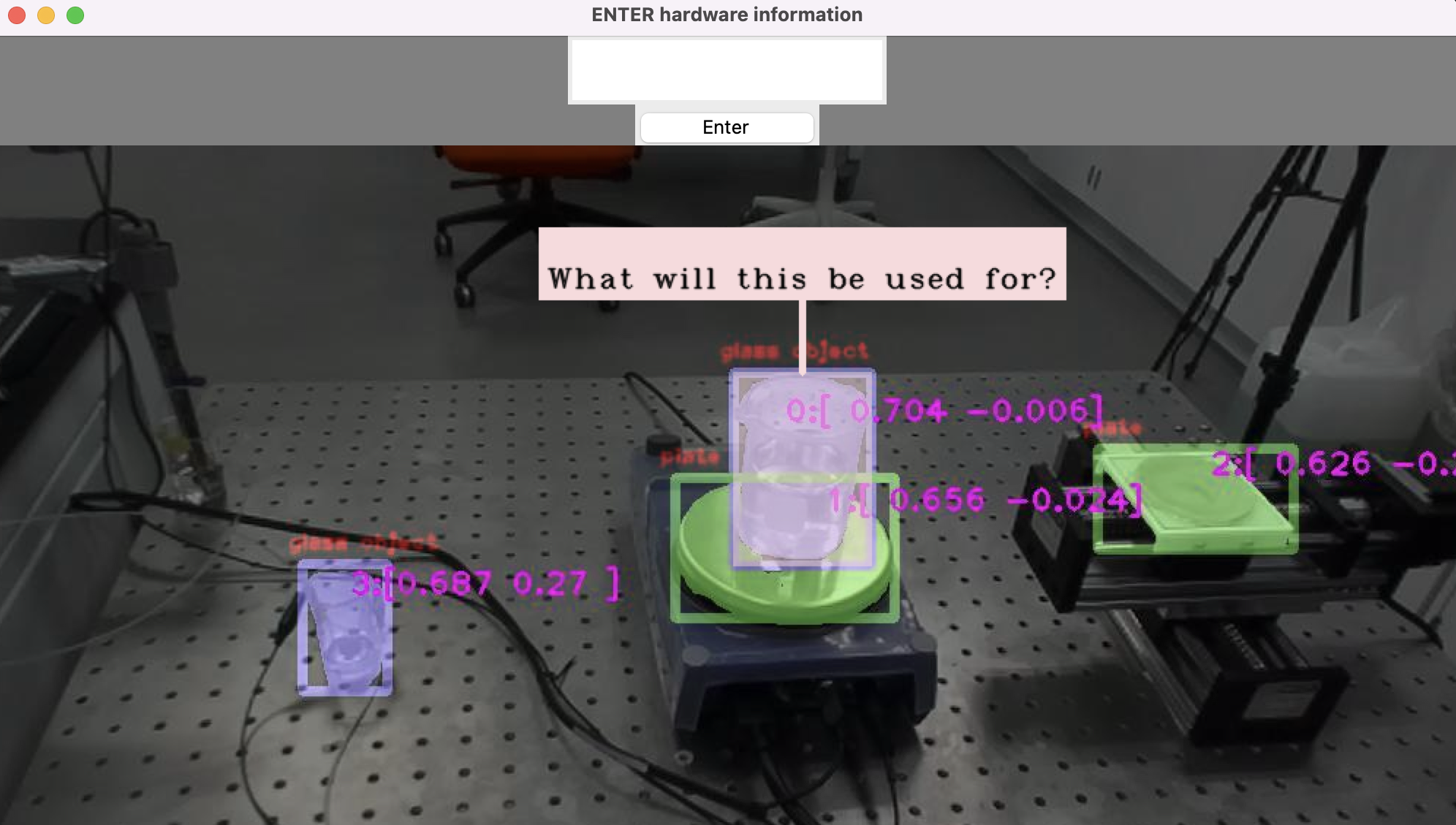}
    \caption{\textbf{Visualization of the user interface for grounding perception.} Users can respond by typing in the text box or by talking to \organa. }
    \label{fig:ground_perception_gui}
    \addcontentsline{toc}{figure}{\cref{fig:ground_perception_gui}}
\end{figure}

\paragraph{Phase 2: Physical Grounding}
If the experiment category is valid, \nlp enters the grounding phase, where \nlp shows the users vessels that it detected using the perception pipeline and the user is asked to assign semantic meaning to the vessels so that \organa can assign each vessel a role in the experiment. This is done using a graphical user interface (see Fig. ~\ref{fig:ground_perception_gui}).  
 \begin{lstlisting}[language=Python]
grounding_question = """
What will this be used for?
""" \end{lstlisting}

The user enters an answer in natural text for each vessel, which is assigned to the vessel. \nlp is called to match the name of the vessel that the user entered to a list of names it has in its knowledge base: 

 \begin{lstlisting}[language=Python]
grounding_perception_prompt="""
These are the official names of hardware in a chemistry lab:
  {0}

These are user descriptions of those hardware, not necessarily that order:
  {1}

Match the official names to the user descriptions using the following format:
[start matching]
<user description>|<official_name>
[end matching]
""" \end{lstlisting}

In the above prompt, \texttt{\{0\}} refers to the list of vessel in the knowledge base and \texttt{\{1\}} refers to the names input by a human. A sample grounding prompt is:

 \begin{lstlisting}[language=Python]
grounding_perception_prompt="""
These are the official names of hardware in a chemistry lab:
  ["washing_beaker", "experiment_beaker", "pH_beaker"]

These are user descriptions of those hardware, not necessarily that order:
  ["main exp beaker", "this beaker is used for washing", "beaker for measuring pH"]

Match the official names to the user descriptions using the following format:
[start matching]
<"this beaker is used for washing">|<"washing_beaker">
<"main exp beaker">|<"experiment_beaker">
<"beaker for measuring pH">|<"pH_beaker">
[end matching]
""" 
\end{lstlisting}

\paragraph{Phase 3: Experiment}

To generate an experiment plan, \nlp is given the following prompt: 

 \begin{lstlisting}[language=Python]
experiment_prompt="""
Experiment goal: {0}
Lab equipment and reagents: {1}

I will tell you how to run a single experiment:

<Thought>{2}</Thought>
<Action>{3}</Action>
<Expected Observation>{4}</Expected Observation>

These are past experiments that you did:

{5}

Can you propose the next experiment to try, as well as the output you expect to see?

This was feedback receieved during the last experiment, incorporate it when planning your next experiment:

<Human Feedback>{6}</Human Feedback>

Rules:
1. Format the plan using <Thought>, <Action>, <Expected Output> tags.
2. If you see feedback in <System Feedback> tags, consider them when planning the next experiment.
"""
\end{lstlisting}

\texttt{\{0\}} is the experiment goal taken from the \texttt{init\_conditions\_dict[`goal']}. \texttt{\{1\}} is the list of lab equipment from \texttt{init\_conditions\_dict[`setup']}. \texttt{\{2-4\}} are taken from the example provided by the user in \texttt{init\_conditions\_dict[`thought']}, \texttt{init\_conditions\_dict[`action']}, \texttt{init\_conditions\_dict[`observation']}. After the first experiment, \texttt{\{5\}} is a memory prompt of past experiments performed. This is important so that\\ \nlp does not propose the same experiment multiple times and is further explained in the ``Memories of past experiments'' section. \texttt{\{6\}} is any feedback returned from users about the experiment. Users are asked for feedback if the results of the experiment do not match the expectations. This is further described in ``Examples of ambiguity and uncertainty mitigation'' below. 

\paragraph{Memories of past experiments} To reduce the number of tokens used for memories, past observations are summarized for every $k$ experiment. We use $k=3$. The  memory prompt is structured as follows:

 \begin{lstlisting}[language=Python]
memory_prompt="""
These are past experiments that you did:

This is a summary of what happened during the first three experiments:

{0}

This is a summary of what happened during the next three experiments:

{1}

...

This is what happened during the most recent experiments:

<Thought>prev_exp[`thought']</Thought>
<Action>prev_exp[`action']</Action>
<Observation>prev_exp[`observation']</Observation>
"""
\end{lstlisting}

Every three experiments, memories are summarized using the following prompt: 

 \begin{lstlisting}[language=Python]
memory_summary="""
These are experiments that a robot did:

{0}

Can you provide a summary of these experiments?
"""
\end{lstlisting}

\looseness=-1
Once \nlp generates a new experiment plan, it is converted to XDL using CLAIRify. As in the original CLAIRify work, we use OpenAI's GPT-3.5 Turbo (\texttt{gpt-3.5-turbo\footnote{\url{https://platform.openai.com/docs/models/gpt-3-5-turbo}})} to generate XDL code from natural language instructions~\cite{AuRo2023Clairify}. The robot performs the experiment and the results are sent back to \nlp so that is can compare them with expected results and plan subsequent experiments. Once all the experiments are completed, \nlp is asked to generate a single final summary for the report: 

 \begin{lstlisting}[language=Python]
final_summary="""
Below are summaries from a series of experiments the robot did:

{0}

Provide a single summary for all these experiments.
"""
\end{lstlisting}

\newpage
\notelabel{appendix:nlp:examples-ambiguity-uncertainty}{Examples of ambiguity and uncertainty mitigation}

\nlp is prompted to rationalize about whether the outcomes of the experiment match user expectations. This is done using the following prompt: 

 \begin{lstlisting}[language=Python]
rationalize_prompt="""
This was an experiment a robot was asked to do:

{0}

After performing the experiment, these are the outcomes:

{1}

Do the actual outcomes of the experiment match the expected outcomes? Begin your answer with <YES> or <NO>, followed by the explanation.
"""
\end{lstlisting}

If \nlp cannot make sense of the experiment outcomes, the user is pinged to determine if there are any issues in the environment. Below is an example of an instance where the experiment results did not match the user expectations:

\begin{small}
\begin{lstlisting}[language=Python]
# rationalization_output
"""
<NO>  The actual outcomes of the experiment do not match the expected outcomes. The pH measured at 5.96 is close to the expected pH of 6, which is within the acceptable range. However, the potential at this pH level is -0.492, which is not more negative than the previous measurement at pH 4.83 (-0.681). This contradicts the expected trend of the potential becoming more negative as the pH increases.

pinging 
human......

Observations don't make sense. Any feedback on the experiment?
"""
\end{lstlisting}
\end{small}

The user then provides feedback on what (if anything) explains the results. \\ \nlp can then incorporate that feedback into the next plan (for example, the user might say `I'm not sure what happened, but repeat the previous experiment just in case', `Nothing is wrong, carry on', `The pump was stuck, it's ok again').

\newpage

\notelabel{appendix:user-study}{User Study}
\addcontentsline{toc}{subsection}{\cref{note:appendix:user-study}}
\paragraph{User Study Participants}
We recruited 8 chemists to test \organa. \newsuggest{}{ These volunteers were from the Department of Chemistry at the University of Toronto.} The expertise of the chemists ranged from novice (little experimental background) to expert (13 years of experience). 25\% of participants identified as female and 75\% as male. Ages ranged from early twenties to early thirties. \update{}{None of the participants were \organa team members or previously familiar with \organa.}

\paragraph{Experiment protocol}Every participant received the same training for the experiment. When they entered the lab, the same member from the \organa team first asked them to fill out the pre-experiment questions (see \textit{Custom user study questionnaire} below). Then, the \organa team member demonstrated how to perform one full run of the manual experiment. These were the steps of the protocol: 
\begin{itemize}
\item Polish a carbon electron on a polishing pad with alumina powder and rinse it with water.
\item To an empty beaker, add 2 mL of 0.5 M pH X buffer solution (X is an integer between 4 to 10), 1 mL of 1 M NaCl solution,  5 mL of water, 2 mL of 10 mM quinone solution using a 10 mL micropipette.
\item Connect reference, working, and ground electrodes to alligator clips inside the experiment beaker, which are already connected to a potentiostat. Press ``play'' on the potentiostat to generate a CV curve. 
\item Remove the electrodes and measure the pH of the solution in the beaker using a pH probe. Rinse the beaker with deionized water.
\end{itemize}
The participant was instructed to repeat these steps for 3 runs (with 3 randomly generated pH values provided by the \organa team member). The \organa team member was always present to answer any questions and make sure the participant was performing the experiment correctly. After they finished the manual experiment, the participant was asked to repeat the experiment, this time by interacting with CLAIRify and \organa in random orders. We show the scripts we provided for interacting with \organa and CLAIRify below. They then filled out the remaining questionnaires, which we also show below.

\begin{lstlisting}[escapechar=@]
@\textbf{CLAIRify script}@

#Experiment 1
Add 2 mL of quinone, 17 mL of NaCl solution, and 1 mL of pH 4 buffer solution. Stir for 10 seconds. Measure the pH and run a CV plot.

#Experiment 2
Add 2 mL of quinone, 17 mL of NaCl solution, and 1 mL of pH 5 buffer solution. Stir for 10 seconds. Measure the pH and run a CV plot.

#Experiment 3
Add 2 mL of quinone, 17 mL of NaCl solution, and 1 mL of pH 6 buffer solution. Stir for 10 seconds. Measure the pH and run a CV plot.

#Experiment 4
Add 2 mL of quinone, 17 mL of NaCl solution, and 1 mL of pH 7 buffer solution. Stir for 10 seconds. Measure the pH and run a CV plot.

#Experiment 5
Add 2 mL of quinone, 17 mL of NaCl solution, and 1 mL of pH 8 buffer solution. Stir for 10 seconds. Measure the pH and run a CV plot.

#Experiment 6
Add 2 mL of quinone, 17 mL of NaCl solution, and 1 mL of pH 9 buffer solution. Stir for 10 seconds. Measure the pH and run a CV plot.
\end{lstlisting}

\begin{lstlisting}[escapechar=@]
@\textbf{ORGANA - start-up phase script}@

>ORGANA
Hello, my name is ORGANA. I am a robot chemist. Is there an experiment I can help you perform today?

>User
I want to run an experiment.

>ORGANA
Could you please provide more details about the experiment you want to run?

>User
I want to create a Pourbaix plot of a quinone, Sodium Anthraquinone-1-sulfonate.

>ORGANA
What is the setup of the lab environment? What hardware and reagents are present?

>User
I have a 2mM solution of the quinone. These are the other solutions I have access to: acetic acid and sodium hydroxide. I also have NaCl. There are three beakers, a pH probe, a potentiostat.

>ORGANA
Can you provide an example of how to run the experiment? 

>User
Add 2 mL of quinone, 17 mL of NaCl solution, and 1 mL of acetic acid. Stir for 10 seconds. Measure the pH and run a CV plot.

>ORGANA
What is the rationale behind this experiment? 

>User
I want to make a solution that has a lower pH than the current solution. Because it's the first scan, I will first measure the pH and run a CV plot.

>ORGANA
What is the expected outcome of the example experiment? 

>User
The pH of the solution is expected to decrease and the redox potential to increase.

>ORGANA
How many experiments are we planning to run in total? 

>User
6

>ORGANA
Ok, thank you! I will start the experiment and let you know if I have any questions.

\end{lstlisting}

\paragraph{System usability scale (SUS) questionnaire}
The SUS questionnaire was given to users both after performing the manual experiment and after using \organa~\cite{sus}. The SUS consists of the following statements, which users rated on a Likert scale from 1 (strongly disagree) to 5 (strongly agree):

\begin{enumerate}
    \item I think that I would like to use this system frequently. 
    \item I found the system unnecessarily complex.
    \item I thought the system was easy to use.
    \item  I think that I would need the support of a technical person to be able to use this system.
    \item I found the various functions in this system were well integrated.
    \item I thought there was too much inconsistency in this system.
    \item I would imagine that most people would learn to use this system very quickly.
    \item I found the system very cumbersome to use.
    \item I felt very confident using the system.
    \item I needed to learn a lot of things before I could get going with this system.
\end{enumerate}

User ratings were transformed to a score out of 10 for each question, where higher is better. The formula for the score $s_i$ for each question $x_i$ is:  

\begin{itemize}
\item $s_i = 2.5*(x_i-1)$ for $x_i \in \{1,3,5,9\}$ 
\item $s_i = 2.5*(5 - x_i)$ for  $x_i \in \{2,4,6,8,10\}$ 
\end{itemize}

\paragraph{NASA Task Load Index (NASA-TLX) questionnaire}
We gave users the NASA-TLX quesetionnaire~\cite{hart2006nasa} both after performing the electrochemistry experiment manually and after using \organa to perform the experiment. The questions are listed below, and the user grades each from 0-20. The score is divided by two to get a score out of 10.

\begin{enumerate}
    \item Mental demand: How mentally demanding was the task? (0: very low, 20: very high)
    \item Physical demand: How physically demanding was the task? (0: very low, 20: very high)
    \item Temporal demand: How hurried or rushed was the pace of the task? (0: very low, 20: very high)
    \item Performance: How successful were you in accomplishing what you were asked to do? (0: perfect, 20: failure)
    \item Effort: How hard did you have to work to accomplish your level of performance? (0: very low, 20: very high)
    \item Frustration: How insecure, discouraged, irritated, stressed, and annoyed were you? (0: very low, 20: very high)
\end{enumerate}
\paragraph{Custom user study questionnaire} We developed an additional custom set of questions for participants before and after each experiment. We show the full questionnaire below:

\begin{lstlisting}[escapechar=@]
@\textbf{Testing ORGANA - User interaction survey}@

@\textbf{Question types}@

@$\medstar$@ Task efficiency: task time, productive time ratio, unnecessary actions
@$\odot$@ Situational awareness: does the user understand what the robot is doing/planning?
@$\smileeq$@ Workload: amount of work loaded on an individual, incl. time pressure, effort exerted, and success

@\textbf{Pre-experiment questions}@

1. @$\smileeq$@ In the lab, some of the tasks I do can be automated.
@$\square$@ Strongly agree
@$\square$@ Agree
@$\square$@ Neither agree nor disagree
@$\square$@ Disagree
@$\square$@ Strongly disagree 

If you wish, please elaborate on what tasks specifically you would wish were automated:


2. @$\medstar$@ I waste a significant amount of time doing repetitive tasks in the lab.
@$\square$@ Strongly agree
@$\square$@ Agree
@$\square$@ Neither agree nor disagree
@$\square$@ Disagree
@$\square$@ Strongly disagree 

3. @$\medstar$@ I find that that every task or most tasks I do in the lab aren't tedious.
@$\square$@ Strongly agree
@$\square$@ Agree
@$\square$@ Neither agree nor disagree
@$\square$@ Disagree
@$\square$@ Strongly disagree 

4. @$\smileeq$@ I would like it if autonomous agents like robots were to take over some of the tasks I need to do in a lab.
@$\square$@ Strongly agree
@$\square$@ Agree
@$\square$@ Neither agree nor disagree
@$\square$@ Disagree
@$\square$@ Strongly disagree 

5. @$\smileeq$@ I would be uncomfortable with an autonomous agent like a robot performing some of the tasks I need to do in the lab.
@$\square$@ Strongly agree
@$\square$@ Agree
@$\square$@ Neither agree nor disagree
@$\square$@ Disagree
@$\square$@ Strongly disagree 

6. @$\odot$@ I would be comfortable interfacing with a robot using a graphic user interface (GUI).
@$\square$@ Strongly agree
@$\square$@ Agree
@$\square$@ Neither agree nor disagree
@$\square$@ Disagree
@$\square$@ Strongly disagree

7. @$\odot$@ I would be comfortable interfacing with a robot using a command-line interface (CLI).
@$\square$@ Strongly agree
@$\square$@ Agree
@$\square$@ Neither agree nor disagree
@$\square$@ Disagree
@$\square$@ Strongly disagree

8. @$\odot$@ I would be comfortable interfacing with a robot using voice commands.
@$\square$@ Strongly agree
@$\square$@ Agree
@$\square$@ Neither agree nor disagree
@$\square$@ Disagree
@$\square$@ Strongly disagree

9. @$\odot$@ I am very familiar with chemistry theory.
@$\square$@ Strongly agree
@$\square$@ Agree
@$\square$@ Neither agree nor disagree
@$\square$@ Disagree
@$\square$@ Strongly disagree

10. @$\odot$@ I have done chemistry experiments before in a lab setting.
@$\square$@ Strongly agree
@$\square$@ Agree
@$\square$@ Neither agree nor disagree
@$\square$@ Disagree
@$\square$@ Strongly disagree

How long have you been doing chemistry experiments for? (indicate approximate number of years)

What is your position? E.g. post-doctoral fellow, graduate student, undergraduate student

Any additional comments?

@\textbf{CLAIRify - experiment}@

@\textit{Task description:}@ You are asked to interact with CLAIRify, a model that can generate robot code from natural language descriptions. You will be provided with a script describing each experiment you would like the robot to perform. When prompted, please type in the first experiment and press Enter; the robot will "run" the experiment and prompt you when it is finished, and then you can enter the next experiment. Please repeat this until all experiments are finished. Once you are done this part and the ORGANA experiments, you will be asked a series of questions.

@\textbf{ORGANA - start-up phase}@

@\textit{Task description:}@ You are asked to interact with a robot chemistry assistant, ORGANA. You will be provided with a script detailing the experiment. The goal of the assistant is to find out information about the experiment, and so it will ask you a series of questions. When you're ready, ORGANA will ask you a series of questions about the experiment. The answers to the questions are provided in the script. You will first respond by typing. Then, you will repeat the process, but this time respond by speaking. Once you are done, please answer the questions below:

11. @$\smileeq$@ I found it easy to interact with ORGANA, either through speaking or typing.
@$\square$@ Strongly agree
@$\square$@ Agree
@$\square$@ Neither agree nor disagree
@$\square$@ Disagree
@$\square$@ Strongly disagree

12. @$\medstar$@  ORGANA was able to capture the experiment I wanted to perform with little to no issues.
@$\square$@ Strongly agree
@$\square$@ Agree
@$\square$@ Neither agree nor disagree
@$\square$@ Disagree
@$\square$@ Strongly disagree

13. @$\medstar$@  I preferred speaking with ORGANA rather than typing.
@$\square$@ Strongly agree
@$\square$@ Agree
@$\square$@ Neither agree nor disagree
@$\square$@ Disagree
@$\square$@ Strongly disagree

14. @$\medstar$@  I believe ORGANA could save me time compared to creating an initial system setup from code.
@$\square$@ Strongly agree
@$\square$@ Agree
@$\square$@ Neither agree nor disagree
@$\square$@ Disagree
@$\square$@ Strongly disagree

Additional comments?

@\textbf{ORGANA - experiment}@

@\textit{Task description:}@ After initialising the system, now ORGANA will run the experiments (6 in total). We will run the experiments in two modes: one where ORGANA will plan the experiments and execute them automatically, only calling the human where results don't make sense, and one where she will wait for user input for each experiment. 

Note: we won't actually be running the experiment, we will call values from a previous experiment in the backend. For one of the experiments, we will purposely repeat the value from the last experiment  in order to "ping" the human due to experimental inconsistency. You will be asked to "look into" this experiment and inform ORGANA on next steps. Once you are done going through both this exercise and the CLAIRify exercise, please answer the questions below:

15. @$\smileeq$@ I found it easy to interact with ORGANA, either through speaking or typing.
@$\square$@ Strongly agree
@$\square$@ Agree
@$\square$@ Neither agree nor disagree
@$\square$@ Disagree
@$\square$@ Strongly disagree

16. @$\medstar$@ I found it less time consuming for Organa to plan the experiments vs typing them in myself each time.
@$\square$@ Strongly agree
@$\square$@ Agree
@$\square$@ Neither agree nor disagree
@$\square$@ Disagree
@$\square$@ Strongly disagree

17. @$\medstar$@ I prefer to input each experiment myself vs having ORGANA plan them herself.
@$\square$@ Strongly agree
@$\square$@ Agree
@$\square$@ Neither agree nor disagree
@$\square$@ Disagree
@$\square$@ Strongly disagree

18. @$\odot$@ I trust Organa to plan and execute the experiments from start to finish.
@$\square$@ Strongly agree
@$\square$@ Agree
@$\square$@ Neither agree nor disagree
@$\square$@ Disagree
@$\square$@ Strongly disagree

19. @$\odot$@  I would like to be kept in-the-loop of the experiment while it is running.
@$\square$@ Strongly agree
@$\square$@ Agree
@$\square$@ Neither agree nor disagree
@$\square$@ Disagree
@$\square$@ Strongly disagree

20. @$\odot$@   I would prefer to be called more often to verify that the experiment running is correct.
@$\square$@ Strongly agree
@$\square$@ Agree
@$\square$@ Neither agree nor disagree
@$\square$@ Disagree
@$\square$@ Strongly disagree

Additional comments?

@\textbf{Post-experiment summary}@

21. @$\smileeq$@ I find having a post-experiment summary valuable.
@$\square$@ Strongly agree
@$\square$@ Agree
@$\square$@ Neither agree nor disagree
@$\square$@ Disagree
@$\square$@ Strongly disagree

22. @$\smileeq$@  I am satisfied with the information that is contained in the post experiment summary.
@$\square$@ Strongly agree
@$\square$@ Agree
@$\square$@ Neither agree nor disagree
@$\square$@ Disagree
@$\square$@ Strongly disagree

23. @$\smileeq$@  I found  the visualisations in the post-experiment summary to be useful and informative.
@$\square$@ Strongly agree
@$\square$@ Agree
@$\square$@ Neither agree nor disagree
@$\square$@ Disagree
@$\square$@ Strongly disagree

24. @$\smileeq$@ I would prefer having more information in the post experiment summary.
@$\square$@ Strongly agree
@$\square$@ Agree
@$\square$@ Neither agree nor disagree
@$\square$@ Disagree
@$\square$@ Strongly disagree

Additional comments?

\end{lstlisting}

\newpage
\notelabel{appendix:hardware-perception}{Hardware for Perception}
\addcontentsline{toc}{subsection}{\cref{note:appendix:hardware-perception}}

We integrated diverse hardware components for conducting a range of chemistry measurements. Specifically, an IKA RET control-visc served as a weighing scale and was utilized for pH measurements by interfacing with a pH probe (Orion ROSS Ultra Refillable pH/ATC Triode Combination Electrodes for Orion Series Meters, Thermo Fisher Scientific). Additionally, a Sartorius BCA2202-1S Entris functioned as a high-precision scale. Cyclic voltammetry scans of solutions were performed using a portable potentiostat~\cite{garcia2023potentiostat} to measure electrochemical properties, which we describe in further detail below. Two cameras were used in our experimental setup.
The Intel RealSense D435i stereo camera, mounted on the robot end-effector, estimated object poses in the scene by detecting fiducial markers through the AprilTag library\cite{olson2011apriltag} and also facilitated the measurement of solution turbidity. Given the prevalence of transparent objects in the laboratory, as highlighted in~\citet{jiang2023robotic}, we employed a ZED Mini camera for transparent object detection and depth estimation~\cite{zedmini}. We utilized the ZED camera depth map in neural depth mode.
The hand-eye calibration for both cameras was executed using an infrastructure robotics library\footnote{\href{https://github.com/uoft-cs-robotics/robot_system_tools}{https://github.com/uoft-cs-robotics/robot\_system\_tools}}. 
These devices were connected to the robot controller machine via USB protocol.

\paragraph{Redox potential feedback}
We ran a cyclic voltammetry (CV) measurement~\cite{elgrishi2018practical} to measure the redox potential of the quinone solution at a given pH.
A portable potentiostat~\cite{garcia2023potentiostat} was connected to the robot controller machine and the CV measurement was conducted with a standard 3-electrode system, where the working electrode was a glassy carbon, and the counter and reference electrodes were silver wires.
The reduction and oxidation peaks in the CV plot were automatically detected by picking up the minimum and maximum values in the measured voltage range.
The mean voltage of the two peak voltages was used as the redox potential of the sample.

\newpage
\notelabel{appendix:hardware-skills} {Hardware for Actions}
\addcontentsline{toc}{subsection}{\cref{note:appendix:hardware-skills}}

For solution heating and stirring, we employed the IKA RET control-visc, which interfaced with the robot workstation using PyVISA~\cite{grecco2023pyvisa}.
For precise liquid transfers between different stations in the electrochemistry experiment, we used a Cavro XCalibur Pump containing a 12-port ceramic valve (Tecan Systems), which transported reagents into the specified container upon request.
For all experiments, we used a Franka Emika Panda arm
robot equipped with a Robotiq 2F-85 gripper.
To facilitate grasping objects from the side in tabletop scenarios, we positioned the end-effector parallel to the ground on the robot's last link.
This configuration was obtained either through a fixed linkage or by incorporating a Dynamixel XM540-W150 servo motor as an additional degree of freedom.

\newpage
\notelabel{appendix:parameter-estimaiton}{Electrochemistry Parameter Estimation}
\addcontentsline{toc}{subsection}{\cref{note:appendix:parameter-estimaiton}}

In the electrochemistry experiment, we had two goals for parameter estimation. The first was to produce a maximum likelihood estimation (MLE) for the model parameters given the data, which we returned at the end of the experiment. The second was to estimate the posterior distribution in the parameter space given the data and individual marginal distribution for each parameter, which we used to give a visual representation of the progress of the experiment to the chemist.

\textbf{Maximum likelihood estimation.} We assume that our data points come from a Gaussian distribution with a mean ($\mu_{eV}$) predicted by the model. We assume that the variance ($\sigma_{eV}$) does not depend on pH, but do not assume that it is known and so we add it as an additional parameter to the estimation. Our model parameters are therefore denoted with:

\begin{align}
    \theta &= (\mathrm{pK_{a1}}, \mathrm{pK_{a2}}, k, E_{\mathrm{inf}}, \sigma_{\mathrm{eV}}) \\
    \mu_{\mathrm{eV}}(\mathrm{pH}) &=
    \begin{cases}
      E_{\mathrm{inf}} -  k (\mathrm{pK_{a2}} - \mathrm{pK_{a1}}) - 2 k(\mathrm{pK_{a1}}  - \mathrm{pH}),  &  \textrm{for }
     \mathrm{pH} < \mathrm{pK_{a1}}\\
      E_{\mathrm{inf}} - k (\mathrm{pK_{a2}} - \mathrm{pH}),   &  \textrm{for } \mathrm{pK_{a1}} \le \mathrm{pH} \le  \mathrm{pK_{a2}}\\
      E_{\mathrm{inf}}, &  \textrm{for } \mathrm{pK_{a2}} <  \mathrm{pH}
    \end{cases} \\
    \sigma_{\mathrm{eV}}(\mathrm{pH}) &= \sigma_{\mathrm{eV}} \\
    \mathrm{eV}(\mathrm{pH}) &\sim \mathcal{N}(\mu_{\mathrm{eV}},\,\sigma_\mathrm{eV}^{2})
\end{align}

where $\mathrm{pK_{a1}}$ and $\mathrm{pK_{a2}}$ are the predicted $\mathrm{pK_{a}}$s, $k$ is the slope term of the Pourbaix curve, $E_{\mathrm{inf}}$ is the bias term (y-shift) of the Pourbaix curve, $\sigma_\mathrm{eV}$ is the predicted variance, and $\mu_\mathrm{eV}$ is the mean. The likelihood for a set of data points $\{(\mathrm{pH}, \mathrm{eV})_i\}$ given model parameters is then:
\begin{equation}
\begin{aligned}
\label{eqn:gaus}
    & p(\{(\mathrm{pH}, \mathrm{eV})_i\} \mid \theta) 
    = \prod_i \frac{1}{\sigma_{\mathrm{eV}}^\theta(\mathrm{pH}_i)\sqrt{2\pi}} \exp\left( -\frac{1}{2}\left(\frac{\mathrm{eV}_i-\mu_{\mathrm{eV}}^\theta(\mathrm{pH})}{\sigma_{\mathrm{eV}}^\theta(\mathrm{pH})}\right)^{\!2}\,\right)
\end{aligned}
\end{equation}

We use \lstinline{fmin} method from \lstinline{scipy.optimize} Python package to produce estimated parameter values, which we return as the experiment output ( \cref{note:appendix:report}).

\textbf{Posterior distribution.} We additionally aim to produce an estimate of the entire posterior distribution over the parameter space. In order to do this we assume uniform prior over the parameters. From there we can express the posterior as:
\begin{align}
\label{eqn:bayes}
    p(\theta \mid \{\mathrm{pH}_i, \mathrm{eV}_i\}) =  \frac{p(\{\mathrm{pH}_i, \mathrm{eV}_i\} \mid \theta) p(\theta)}{p(\{\mathrm{pH}_i, \mathrm{eV}_i\})}
\end{align}
With uniform prior, up to a normalization constant we have:
\begin{align}
\label{eqn:posterior}
    p(\theta \mid \{\mathrm{pH}_i, \mathrm{eV}_i\}) &\propto p(\{\mathrm{pH}_i, \mathrm{eV}_i\} \mid \theta) \\
    &\propto \prod_i \frac{1}{\sigma_{\mathrm{eV}}^\theta(\mathrm{pH}_i)\sqrt{2\pi}} \exp\left( -\frac{1}{2}\left(\frac{\mathrm{eV}_i-\mu_{\mathrm{eV}}^\theta(\mathrm{pH})}{\sigma_{\mathrm{eV}}^\theta(\mathrm{pH})}\right)^{\!2}\,\right)
\end{align}

This produces a normalized likelihood function over the space of parameters. Based on this, we can now estimate marginal posterior distribution for each parameter.
We do this using importance sampling, utilizing samples from a uniform distribution covering the possible range for each parameter. We can express the marginal distribution as follows; we demonstrate this for $p(\mathrm{pK}_{\mathrm{a1}})$ but it is the same for all other parameters:
\begin{align}
    \begin{split}
    p(\mathrm{pK}_{\mathrm{a1}} \mid \{\mathrm{pH}_i, \mathrm{eV}_i\}) &= \int_\theta \mathrm{pK}_{\mathrm{a1}}^\theta \cdot p(\theta \mid \{\mathrm{pH}_i, \mathrm{eV}_i\}) d\theta \\
    & \approx \frac{1}{N} \sum_{j=1}^N \frac{p(\theta \mid\{\mathrm{pH}_i, \mathrm{eV}_i\})} {q(\theta)} \mathrm{pK}_{\mathrm{A1}}^\theta \textrm{, where } \theta \sim q \\
    &\approx \frac{1}{S} \sum_{j=1}^N p(\{\mathrm{pH}_i, \mathrm{eV}_i\} \mid \theta) \mathrm{pK}_{\mathrm{a1}}^\theta, \\
    &\quad \textrm{where } \theta \sim Unif \textrm{and $S$ is a normalization constant}
    \end{split}
\end{align}

Finally, we can give the estimate for the distribution of the mean values for possible models at each pH (essentially a distribution over possible model lines). We use this to visually represent our current belief over possible models and give the user an  overview of the current experiment status.
\begin{align}
    \begin{split}
    p(\mu_\mathrm{eV}(\mathrm{pH}) \mid \{\mathrm{pH}_i, \mathrm{eV}_i\})
    &= \int_\theta \mu_\mathrm{eV}(\mathrm{pH} \mid \theta) p(\theta \mid \{\mathrm{pH}_i, \mathrm{eV}_i\}) d\theta \\
    &\approx \frac{1}{N} \sum_{j=1}^N \frac{p(\theta \mid\{\mathrm{pH}_i, \mathrm{eV}_i\})} {q(\theta)} \mu_\mathrm{eV}(\mathrm{pH} \mid \theta), 
    \quad\textrm{where } \theta \sim q\\
    &\approx \frac{1}{S}\sum_{j=1}^N p(\{\mathrm{pH}_i, \mathrm{eV}_i\} \mid \theta)\mu_\mathrm{eV}(\mathrm{pH} \mid \theta), \\
    &\quad \textrm{where } \theta \sim Unif \textrm{and $S$ is a normalization constant}
    \end{split}
\end{align}

\newpage

\notelabel{appendix:PDDL}{Integrated PDDLStream with Scheduling}
\addcontentsline{toc}{subsection}{\cref{note:appendix:PDDL}}

\paragraph{PDDL Language Description}

\revise{}{Planning Domain Definition Language (PDDL) aims to standardize research on AI planning so that planning algorithms are reusable and comparable~\cite{mcdermott1998pddl}. It consists of two parts: the domain definition and the problem definition.
The PDDL domain is designed to model the physics of a domain using predicates and actions. Predicates describe properties of specific types of objects, representing the world state and their relationships; they are either true or false. Actions represent operations and transitions in the world. They take some parameter variables as input, have preconditions that specify when the actions are applicable, and have effects that describe how the world changes after the action is applied.
Here is an example of a PDDL domain for a washing station and a robot in a chemistry lab.}


\begin{lstlisting}[language=lisp, breaklines=true, label=lst:pddldomain, numbers=left, caption={PDDL domain definition for a washing task.}, escapeinside={<*@}{@*>}]

(define (domain washing-domain)
  (:requirements :strips :typing)

  ;; Define the object types
  (:types
      object
      location
      glassware - object
      beaker vial - glassware
      washer
      robot
  )
  
  ;; define constants
  (:constants
      table_loc washing_station_loc - location
  )
  ;; Define predicates
  (:predicates
    (at ?obj - object ?loc - location) ; the object `obj` at the location `loc`
    (is_picked ?obj - object ?rob - robot) ; object `obj` is picked by robot `rob` 
    (is_free ?robot - robot) ; robot `rob` is free
    (is_washed ?glsw - glassware) ; glassware `glsw` is washed (cleaned)
  )
  
  ;; Define actions
  (:action pick
    :parameters (?rob - robot ?obj - object ?loc - location)
    :precondition (and
      (at ?obj ?loc)
      (is_free ?rob)
      (not (is_picked ?obj ?rob))
      )
    :effect (and
      (not (is_free ?rob))
      (not (at ?obj ?loc))
      (is_picked ?obj ?rob)
    )
  )
  (:action place
    :parameters (?rob - robot ?obj - object ?loc - location)
    :precondition (and
      (not (is_free ?rob))
      (is_picked ?obj  ?rob)
      )
    :effect (and
      (is_free ?rob)
      (at ?obj ?loc)
      (not (is_picked ?obj ?rob ) )
    )
  )
  
  (:action wash
    :parameters (?glsw - glassware ?washer - washer )
    :precondition (and
      (not (is_washed ?glsw ))
      (at ?glsw washing_station_loc)
      )
    :effect (and
      (is_washed ?glsw )
      )
  )
)
\end{lstlisting}


\revise{}{In the domain description, the domain name and the requirements are first defined. Then, objects and their types are specified. A type is similar to a class in object-oriented programming, and it can be a subclass of another class. For example, in this domain, both \texttt{beaker} and \texttt{vial} are of type \texttt{glassware}, which, in turn, is of type \texttt{object}.
This domain has two constants associated with it, which are unchanging locations in this lab space.
Next, the predicates of the domain are defined. \texttt{(at ?obj - object ?loc - location)} specifies that the variable \texttt{?obj} of type \texttt{object} is located at the variable \texttt{?loc} of type \texttt{location}. The value of these variables is grounded when the PDDL problem is solved. 
Finally, in this domain, three actions are defined: \texttt{pick}, \texttt{place}, and \texttt{wash}.
For example, the action \texttt{wash} accepts two variables as input, namely, \texttt{?glsw} of type \texttt{glassware} and \texttt{?washer} of type \texttt{washer}. There are two preconditions associated with the \texttt{wash} action that must both be satisfied: the glassware should not be washed to avoid washing the glassware multiple times \texttt{(not (is\_washed ?glsw ))}, and the glassware should be at the washing station 
\texttt{(at ?glsw washing\_station\_loc)}. When the \texttt{wash} action is applied, it will affect the world by setting the state of the glassware to washed, i.e., \texttt{(is\_washed ?glsw)}.}

\begin{lstlisting}[language=lisp, label=lst:pddlproblem, numbers=left, caption={PDDL problem definition for a washing task.}, escapeinside={<*@}{@*>}]

(define (problem washing-problem)
  (:domain washing-domain)

  ;; Define objects
  (:objects
      beaker1 - beaker
      washer - washer 
      Franka - robot
  )

  ;; Define the initial state
  (:init
    (at beaker1 table_loc)
    (is_free Franka)
    )

  ;; Define the goal state
  (:goal
    (and
        (is_washed beaker1)
    )
  )
)
\end{lstlisting}


\revise{}{Snippet~\ref{lst:pddlproblem} provides an example problem associated with the domain in Snippet~\ref{lst:pddldomain}. First, the problem name and the domain related to this problem are defined. Next, the set of objects that exist in the workspace is specified; in this case, we have one beaker, one washer, and one Franka robot. Following that, the initial and final states are defined. Notice that all predicates associated with objects are set to \texttt{False} by default (closed world assumption) unless they are set to \texttt{True} in the initial state. In this problem, the \texttt{beaker1} is initialized at \texttt{table\_loc}, and the \texttt{Franka} arm is free at the beginning. The goal is to have \texttt{beaker1} washed.}

Using the Fast-Downward PDDL solver~\cite{helmert2006fast}, the following plan is returned to solve the problem:

\begin{lstlisting}[language=lisp, label=lst:pddlsoloution, numbers=left, caption={PDDL solution for the washing task.}, escapeinside={<*@}{@*>}]
(pick franka beaker1 table_loc)
(place franka beaker1 washing_station_loc)
(wash beaker1 washer)
\end{lstlisting}

More information about the PDDL language and examples can be found here~\footnote{\url{https://planning.wiki/}}.

\paragraph{PDDLStream with scheduling for concurrent task execution}
\revise{}{
PDDLStream extends PDDL by defining predicates, actions, initial state, goal state, and streams. The stream $S(\textbf{x})$ over a tuple of literals $\textbf{x}$ acts as a conditional sampler, declaring the satisfaction of the relation between its input and output tuples. Unlike PDDL, PDDLStream solves the task and motion planning problem together.
Following this description of the PDDL language, we can define a general action in PDDLStream as follows:}

\begin{lstlisting}[language=lisp, label=lst:pddlstream-action, numbers=left, caption={PDDLStream action defintion.}, escapeinside={<*@}{@*>}]
(:action Action
  :parameters ( <parameters> )
  :precondition (and
    <preconditions>
  )
  :effect (and
    <effects>
  )
)
\end{lstlisting}
\revise{}{In Snippet~\ref{lst:pddlstream-action}, \texttt{<...>} is a placeholder; for example, \texttt{<parameters>} represents the set of action parameters. \texttt{<preconditions>} includes both the set of predicate preconditions and the streams associated with the action, which are evaluated optimistically.}



\revise{}{In \organa, we facilitate scheduling and parallel task execution by converting instantaneous actions $\texttt{action} \in \mathcal{A}$ into durative actions, incorporating starting $\texttt{action-start}$ and ending $\texttt{action-end}$ points as defined in PDDL2.1~\cite{fox2003pddl2}. This transformation involves updating preconditions and effects to align with the requirements and constraints of concurrent planning.
The new requirements pertain to the availability of agents and minimizing the overall plan execution cost, with a focus on reducing the makespan.}

\begin{lstlisting}[language=lisp, label=lst:concurrent-pddlstream, numbers=left, caption={Concurrent PDDLStream action definition.}, escapeinside={<*@}{@*>}]
(:action Action-start
  :parameters (?agent - agent ?t - timing <parameters>)
  :precondition (and
    <preconditions>
    (at_time ?t)
    (is_free ?agent)
  )
  :effect (and 
    (agent_at_time ?agent ?t)
    (not (is_free ?agent))
    ;; returning the current global timing; a time varying cost function
    (increase (total-cost) (cost_start_<Action> ?t))
  )
)

(:action Action-end
  :parameters (?agent - agent ?<agent>_t ?t ?new_t - timing <parameters>)
  :precondition (and
    <preconditions>
    (at_time ?t)
    (not (is_free ?agent))
    (agent_at_time ?agent ?<agent>_t)
    (update_time_<Action> ?<agent>_t ?t ?new_t) ;; stream generator function
  )
  :effect (and
    <effects>
    (not (agent_at_time ?agent ?<agent>_t))
    (is_free ?agent)
    (when (not (= ?t ?new_t))
      (and (not (at_time ?t)) (at_time ?new_t))
    )
    ;; returning the action time duration
    (increase (total-cost)(cost_end_<Action>))
  )
)
\end{lstlisting}


\revise{}{In \texttt{Action-start}, the responsible agent for the action should be free as a precondition. The \texttt{at\_time} predicate keeps track of global execution timing fluent. As an effect of starting the action, the agent will no longer be free, and the time at which the agent starts the action is tracked using \texttt{(agent\_at\_time ?agent ?t)}. Moreover, the cost associated with starting the action is equal to the current global timing and is added to the \texttt{total-cost}. Thus, the cost of \texttt{Action-start} is time-varying, depending on when the action begins, and is determined during the planning search step. By minimizing the total cost, the time at which an action starts is reduced.}


\revise{}{In the preconditions of \texttt{Action-end}, it checks that the predicate \texttt{(at\_time ?t)} associated with the current global timing fluent is \texttt{True}. This is because, while the action was in progress, several other actions by different agents might have started or ended, potentially altering the global timing. In line 23 of Snippet~\ref{lst:concurrent-pddlstream}, \texttt{(update\_time\_<Action> ?<agent>\_t ?t ?new\_t)} is a stream generator that is evaluated eagerly to update the global timing (see Snippet~\ref{lst:update-time-stream}). By applying the effects of \texttt{Action-end}, the agent becomes free, the global timing is updated if necessary, and the \texttt{total-cost} is increased by the duration of the action.}

\begin{lstlisting}[language=python, label=lst:update-time-stream, numbers=left, caption={Concureent PDDLStream: stream generator function to update the global timing.}, escapeinside={<*@}{@*>}]
update_time_<Action>(t_agent, t):
"""
t_agent: the time agent starts the action
t: the current time
T_action: the action duration
t_max: the max time that the problem/task should be solved
"""
  t'_agent = t_agent + T_action
  t' = max(t, t'_agent)
  if not(t_agent <= t <= t'_agent) or t' >= t_max:
    return None
  yield t'
\end{lstlisting}

\revise{}{This section outlines the necessary modifications to PDDLStream for addressing the scheduling problem during planning. However, the exact changes may vary based on the domain definition and specific setup. For instance, one agent performing a part of the experiment may render it infeasible for another agent to execute a different part, as their tasks could interfere with each other. Readers can access the PDDLStream domain, stream, and problem definition for concurrent task execution related to the electrochemistry experiment in the GitHub repository associated with this paper.}

\newpage

\notelabel{appendix:report}{Analysis Report}

\addcontentsline{toc}{subsection}{\cref{note:appendix:report}}

\begin{figure}[H]
    \centering
    \includegraphics[width=0.8\textwidth]{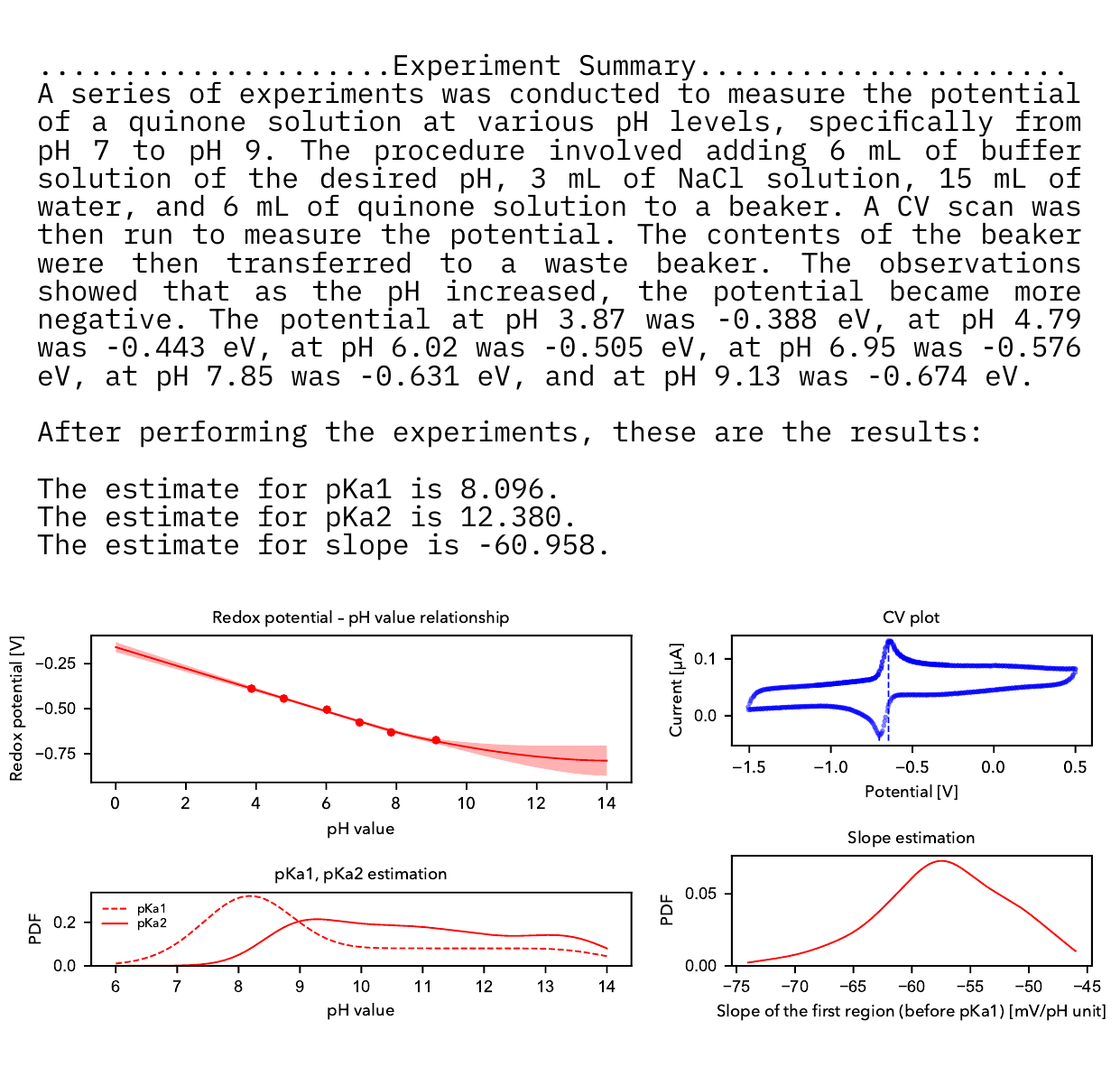}
   \caption{Automatically-generated experiment report: full experiment summary.}
    \label{fig:report_summ}
    \addcontentsline{toc}{figure}{\cref{fig:report_summ}}
\end{figure}

\begin{figure}[H]
    \centering
    \includegraphics[width=0.8\textwidth]{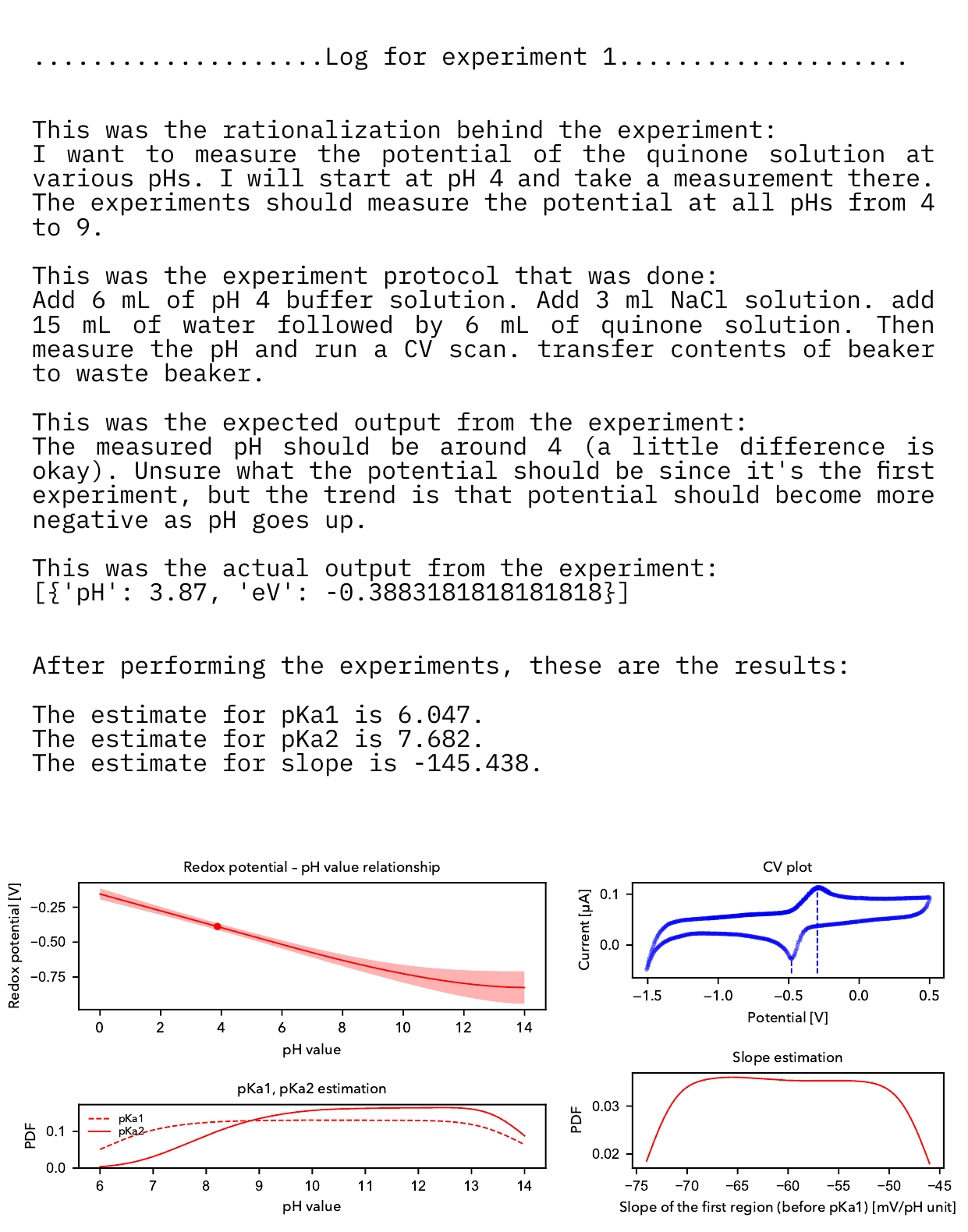}
   \caption{Automatically-generated experiment report: Experiment 1 log.}
    \label{fig:report_1}
    \addcontentsline{toc}{figure}{\cref{fig:report_1}}
\end{figure}

\begin{figure}[H]
    \centering
    \includegraphics[width=0.8\textwidth]{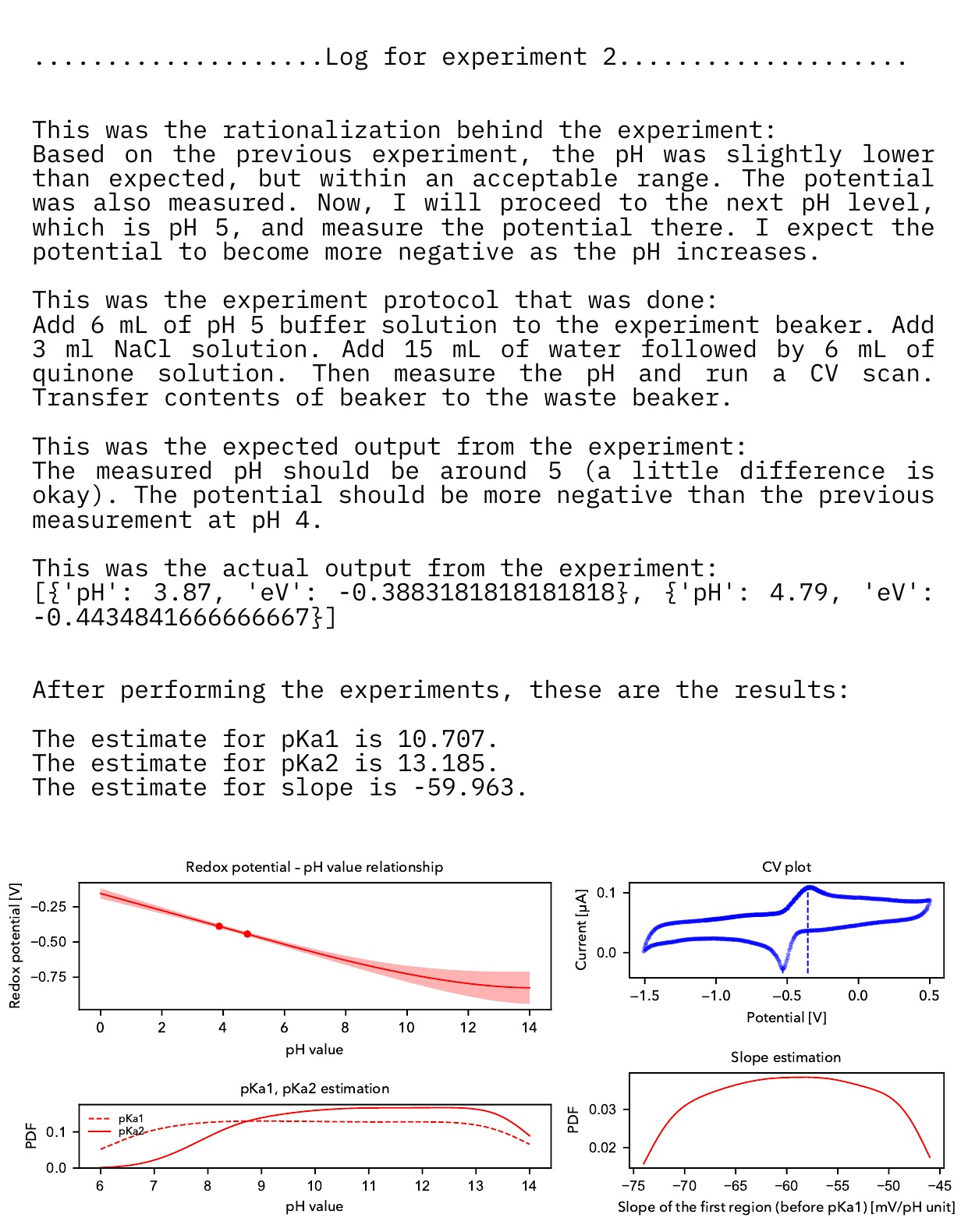}
   \caption{Automatically-generated experiment report: Experiment 2 log.}
    \label{fig:report_2}
    \addcontentsline{toc}{figure}{\cref{fig:report_2}}
\end{figure}

\begin{figure}[H]
    \centering
    \includegraphics[width=0.8\textwidth]{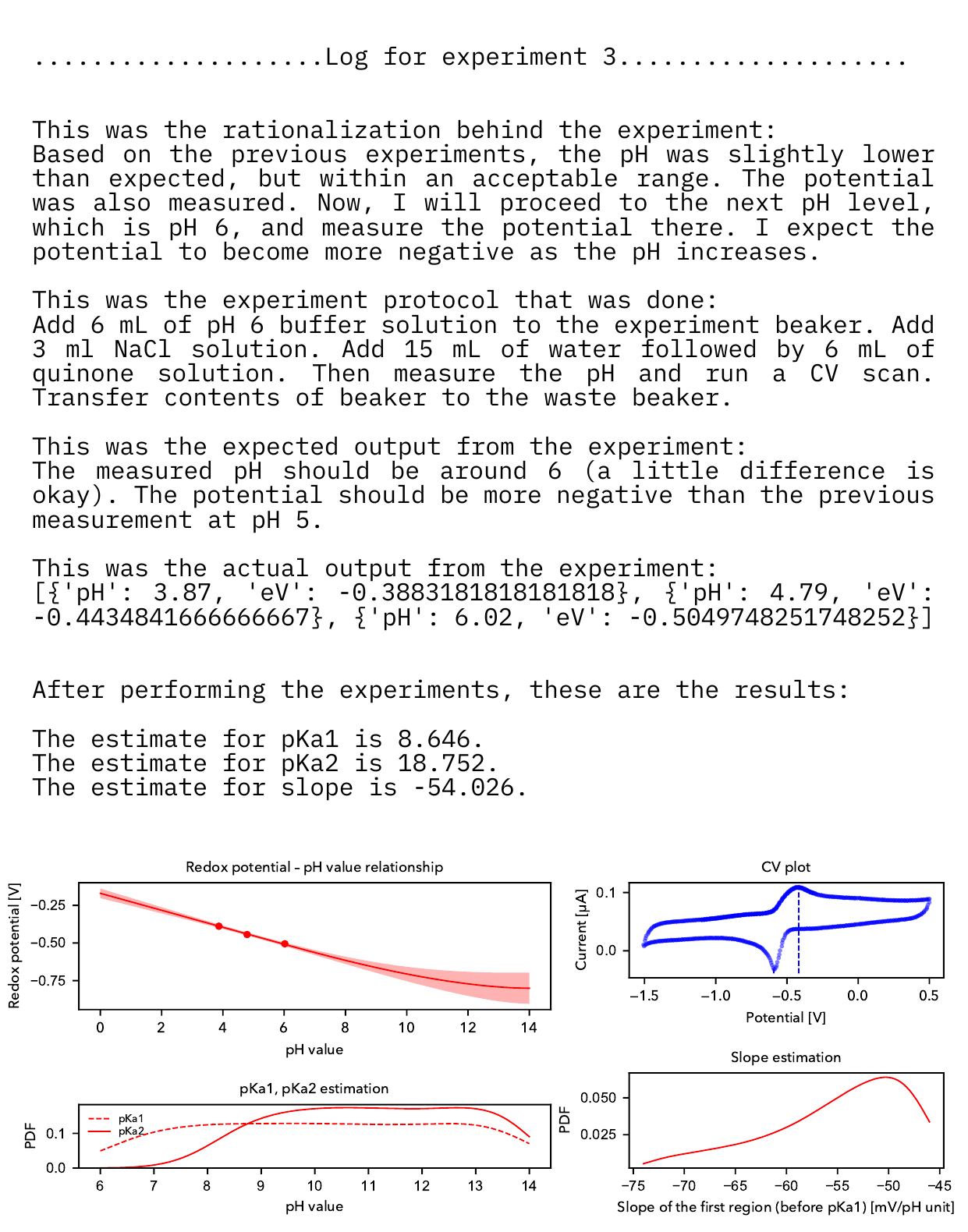}
   \caption{Automatically-generated experiment report: Experiment 3 log.}
    \label{fig:report_3}
    \addcontentsline{toc}{figure}{\cref{fig:report_3}}
\end{figure}

\begin{figure}[H]
    \centering
    \includegraphics[width=0.8\textwidth]{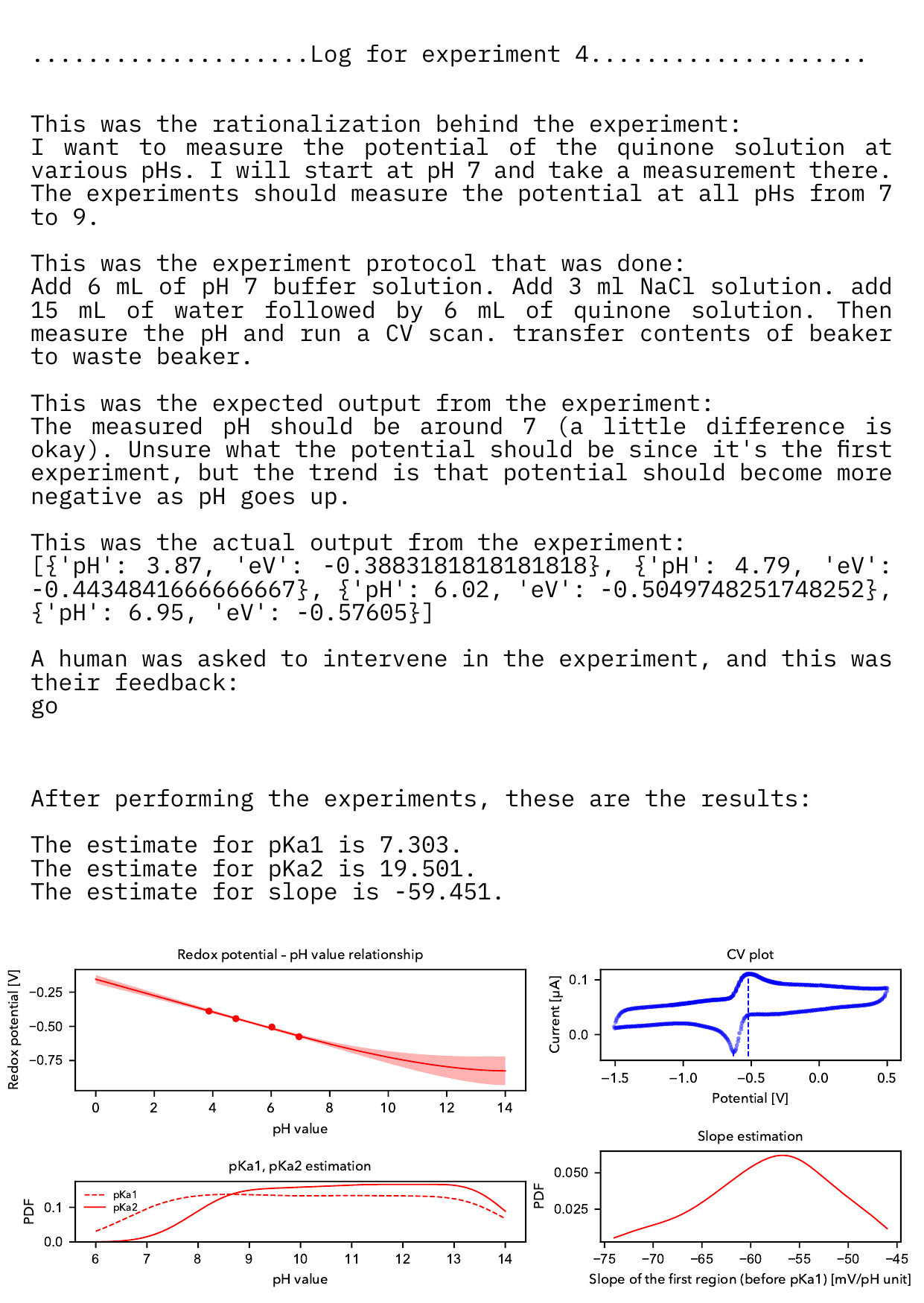}
   \caption{Automatically-generated experiment report: Experiment 4 log.}
    \label{fig:report_4}
    \addcontentsline{toc}{figure}{\cref{fig:report_4}}
\end{figure}

\begin{figure}[H]
    \centering
    \includegraphics[width=0.8\textwidth]{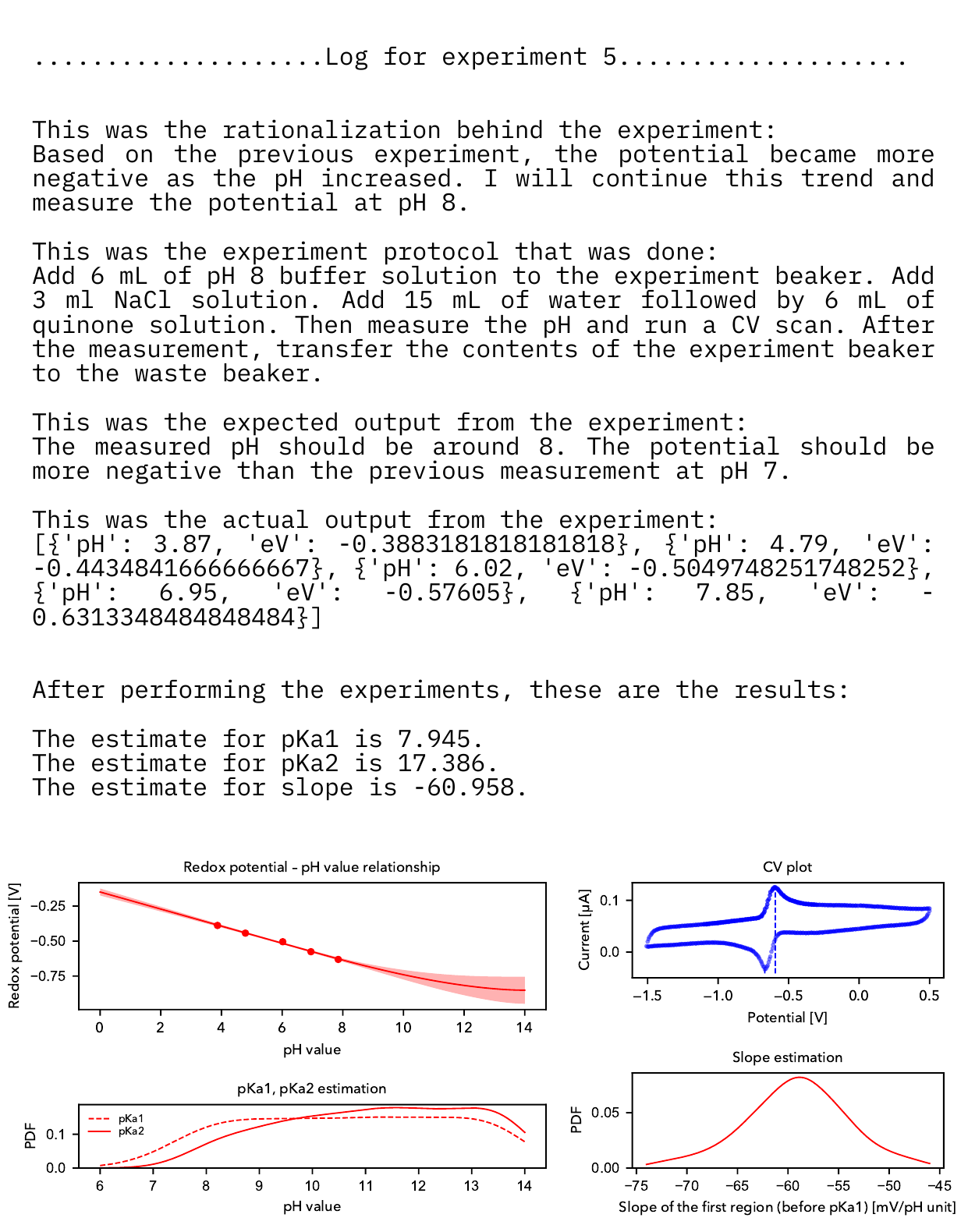}
   \caption{Automatically-generated experiment report: Experiment 5 log.}
    \label{fig:report_5}
    \addcontentsline{toc}{figure}{\cref{fig:report_5}}
\end{figure}

\begin{figure}[H]
    \centering
    \includegraphics[width=0.8\textwidth]{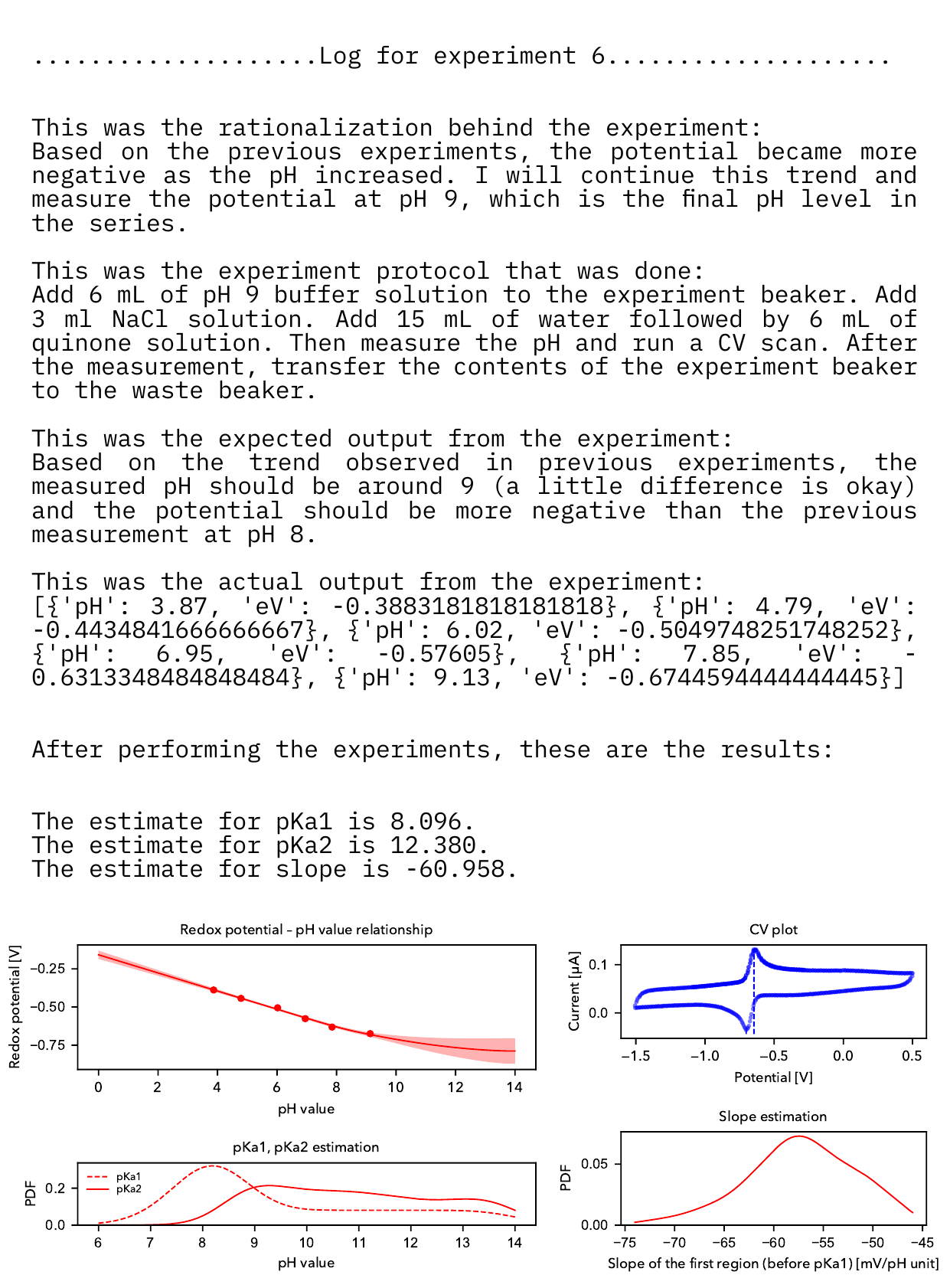}
   \caption{Automatically-generated experiment report: Experiment 6 log.}
    \label{fig:report_6}
    \addcontentsline{toc}{figure}{\cref{fig:report_6}}
\end{figure}

We present an example report generated by the \organa for electrochemistry experiments aimed at obtaining the Pourbaix plot in \cref{fig:report_summ,fig:report_1,fig:report_2,fig:report_3,fig:report_4,fig:report_5,fig:report_6}. \revise{}{The report is from one of the three full runs of the experiment that we report in the paper (note that it is from a different run than the one presented in the main paper).}

\textbf{Plots vs. reported values.}
In \cref{fig:report_summ,fig:report_1,fig:report_2,fig:report_3,fig:report_4,fig:report_5,fig:report_6}, the report consists of experimental logs and summaries, as well as plots representing the marginal distributions for each individual parameter.
Their purpose is to give the chemist an overview of the experiment's current progress, as we cannot show the full distribution over 5 model parameters.
The reported values in this section are the maximum likelihood estimate for the full parameter set in the combined distribution, which may differ from the maximum values in each individual marginal distribution.

\end{document}